\documentclass[10.5pt]{article}

\usepackage{amsmath,amsfonts,amssymb}
\usepackage{mathrsfs}
\usepackage{hyperref}
\usepackage{changes} 
\usepackage{float}
\usepackage[linesnumbered,ruled]{algorithm2e}

\newcommand{\dsp}{\displaystyle}

\newtheorem{theorem}{Theorem}
\newtheorem{proposition}{Proposition}
\newtheorem{lemma}{Lemma}

\def\E{\mathbb{E}}
\def\R{\mathbb{R}} 

\definecolor{SolarizedGreen}{RGB}{133,153,0}
\definecolor{DarkBlue}{RGB}{0,0,200} 
\definecolor{DarkRed}{RGB}{225,0,0} 

\begin{document}

\title{Reinforcement Learning for a Discrete-Time Linear-Quadratic Control Problem with an Application}

\author{Lucky Li\thanks{Thank all the teachers who have given me the knowledge.} \\ \\ 
\normalsize College of Computing, Data Science, and Society, University of California, Berkeley, CA 94720 \\ 
}

\date{}

\maketitle 

\abstract{
We study the discrete-time linear-quadratic (LQ) control model using reinforcement learning (RL). Using entropy to measure the cost of exploration, we prove that the optimal feedback policy for the problem must be Gaussian type. Then, we apply the results of the discrete-time LQ model to solve the discrete-time mean--variance asset-liability management problem and prove our RL algorithm's policy improvement and convergence. Finally, a numerical example sheds light on the theoretical results established using simulations.}

\section{Introduction} 

The concept of reinforcement learning (RL) can be traced back to Minsky (1954), who studied the theory of neural-analog reinforcement systems and its application to the brain model problem. Since then, RL, as a subfield of machine learning, has achieved significant theoretical and technical advancements across various fields, including engineering, biostatistics, economics, business, and financial investment. More recently, RL has shown increasing applicability to real-world problems such as biological data analysis, autonomous driving, robotics control, computer vision, and gaming. Yu et al. (2000) provided an overview of successful RL applications, highlighting its use in adaptive treatment regimes for chronic diseases and critical care, automated clinical diagnosis, and other healthcare domains like clinical resource allocation and optimal process control. They also discussed the challenges, open issues, and future directions for RL research in healthcare.
Wang and Zhou (2019) noted that the application of RL in quantitative finance has attracted more attention in recent years. One of the main reasons is that today's electronic markets provide vast amounts of microstructured data for training and adaptive learning, surpassing the capabilities of human traders and portfolio managers of the past. Considerable research has been conducted in this area. In our work, we focus on a discrete-time linear-quadratic (LQ) control problem, where some parameters of the dynamical system are unknown. We explore how to learn the optimal controller through interaction with the dynamical system. Finally, we apply RL to address a financial problem using the discrete-time LQ framework.

The initial motivation for this work stems from our undergraduate courses in economics, data sciences, and computer sciences. In financial investment, mean--variance analysis is a crucial tool and a core concern for rational economic agents. Markowitz (1952) initially proposed a single-period framework to study the portfolio selection problem, where investors aim to construct a portfolio that maximizes the expected total return for a given level of risk, as measured by variance. Li and Ng (2000) derived the analytical solution to the discrete-time mean-variance portfolio problem, while Zhou and Li (2000) addressed the continuous-time mean--variance problem. Additionally, Li et al. (2017) explored the discrete-time mean--variance asset-liability management problem.
In real-world scenarios, we often only have access to the most recent price data point at each decision-making time, leading to wealth systems with unknown parameters. To address these challenges, Wang and Zhou (2019) tackled the continuous-time mean--variance problem using an entropy-regularized reward function. Cui et al. (2023) developed an RL framework to solve the discrete-time mean--variance model under more general assumptions. Inspired by these works, we aim to study the discrete-time mean--variance asset-liability management problem. Our research focuses on achieving the optimal trade-off between exploration and exploitation by incorporating the entropy-regularized reward function.

In the discrete-time mean--variance asset-liability management problem, liability is a key feature that significantly differentiates it from the model in Cui et al. (2023). This distinction makes the problem more applicable to real-world financial models and also presents greater challenges in terms of solving techniques compared to those addressed by Cui et al. (2023). Fortunately, our studies at Berkeley have positioned us at the forefront of developments in the era of big data. We adopt techniques from the following courses: Math54 (Linear Algebra), EECS16A (Designing Information Devices and Systems I), Data C182 (Designing, Visualizing, and Understanding Deep Neural Networks), CS188 (Introduction to Artificial Intelligence), and CS189 (Introduction to Machine Learning), to tackle the complexities of our problem.
In particular, CS188 and CS189 introduce various algorithms for optimization and statistical learning, which are integral to RL and artificial intelligence. Most RL algorithms consist of two iterative procedures: policy evaluation and policy improvement (see Sutton and Barto (2018)). Policy evaluation provides an estimated value function for the current policy, while policy improvement updates the current policy to enhance the value function. In this work, we prove a policy improvement theorem, which is a crucial prerequisite for interpretable RL algorithms. This theorem ensures that the iterated value functions are non-increasing in terms of a dynamic minimization problem and ultimately converge to the optimal value function of our exploratory mean-variance asset-liability management problem.


RL has been widely applied not only in quantitative finance but also in real-world problems such as biological data analysis, healthcare, autonomous driving, robot control, and computer vision over the past decades. We plan to develop more general RL methods to address real-world problems in future research. Since RL facilitates online learning, allowing iterative refinement of policies with new information, it can better capture complex patterns, thereby increasing both the efficiency and effectiveness of solving real-world problems.
In this work, we consider a dynamic optimization problem within the framework of linear systems. However, most real-world systems are nonlinear. Dogan et al. (2023) studied the learning-based control of nonlinear systems, highlighting a significant theoretical challenge: there may not be explicit controllers for a given set of cost and system parameters in the nonlinear setting. To address this challenge, we plan to further study a control-theoretic analysis using techniques from dynamic optimization theory, extending beyond what is covered in CS188 and CS189. We then aim to pave the way for investigating a policy iteration RL method to tackle real-world problems in future research.


The rest of the paper is organized as follows: Section 2 presents the discrete-time exploratory linear-quadratic (LQ) problem and derives the value function of this problem according to the optimal policy In Section 3, we apply the established results to study the discrete-time mean-variance asset-liability management problem. We then prove the policy improvement and convergence in Section 4. Section 5 illustrates the theoretical results using simulations. Finally, we conclude this work in Section 6.

\section{Formulation} 

Consider the discrete-time stochastic system 
\begin{equation}\label{eq:state} 
\left\{\begin{array}{ll}
x_{t+1} = A x_t + B u_t + (C x_t + D u_t)w_t^x, & \\
y_{t+1} = \bar A y_t + \bar C y_t w_t^y, & \quad \mbox{for } t= 0,1, \cdots, T-1, \\
\end{array}\right.
\end{equation}
where $(x_0,y_0)' \in \mathbb{R}^2$ is the initial state, $(x_t, y_t)' \in \mathbb{R}^2$ is the state, $u_t \in \mathbb{R}^m$ is the control, $u=\{u_t, 0 \leq t<T-1\}$ is the control process, and $w_t^x$ and $w_t^y$ are standard white noises. 
The coefficients $A, C, \bar A, \bar C \in \mathbb{R}$ and $B, D \in \mathbb{R}^{1 \times m}$ are constant. 

Assume that the initial condition $(x_0,y_0)'$ is independent of the noises $w_t^x$ and $w_t^u$, $t = 0, 1, \cdots, T-1$, and all the noises have 
$$
\mathbb{E}[w_t^x] = \mathbb{E}[w_t^y] = 0 
$$
with 
$$
\begin{aligned}
\mathbb{E}[(w_t^x)^2] = \mathbb{E}[(w_t^y)^2] = 1, \quad 
\mathbb{E}[w_t^x w_t^y] = \rho. 
\end{aligned}
$$

We consider the following discrete-time linear-quadratic (LQ) problem 
\begin{align}\label{eq:Vu}
V^{u}(0,x_0,y_0)  = \min_{u} ~ \E\left[\begin{pmatrix}x_T \\ y_T \end{pmatrix}' Q_T \begin{pmatrix}x_T \\ y_T \end{pmatrix} \right]. 
\end{align}
This is a special LQ problem, because we only study the terminal term in the above objective function. We will further develop more general LQ problem in our future research. 

In the reinforcement learning framework, the density function of the control process $u$ 
is denoted by $\pi=\{\pi_t, 0 \leq t<T-1\}$. Its accumulative entropy can be presented by 
$$
\mathcal{H}(\pi):=-\sum_{t=0}^{T-1} \int_{\mathbb{R}^m} \pi_t(u) \ln \pi_t(u) \mathrm{d}u. 
$$ 

The exploratory LQ problem becomes
\begin{align}\label{eq:Vpi}
V^{\pi}(0,x_0,y_0)  = \min_{\pi} ~ \E \left[\begin{pmatrix}x_T \\ y_T \end{pmatrix}' Q_T \begin{pmatrix}x_T \\ y_T \end{pmatrix} +\lambda \sum_{t=0}^{T-1} \int_{\R} \pi_t(u) \ln\pi_t(u) du \right], 
\end{align}
where the temperature parameter $\lambda$ measures the trade-off between exploitation and exploration in this problem. In the following, we will derive the optimal feedback controls and the corresponding optimal value functions of problem (\ref{eq:Vpi}) theoretically. 

Define the value function $J(t, x_t, y_t)$ under any given policy $\pi$: 
\begin{equation}
\begin{aligned}
J(t, x_t, y_t) 
& = \mathbb{E}\left[\begin{pmatrix}x_T \\ y_T \end{pmatrix}' Q_T \begin{pmatrix}x_T \\ y_T \end{pmatrix} 
+ \lambda \sum_{t=0}^{T-1} \int_{\mathbb{R}^m} \pi_t(u) \ln \pi_t(u) \mathrm{d} u\bigg| \mathcal{F}_t \right], 
\end{aligned}
\end{equation}
where $Q_T$ is a symmetric matrix of $\mathbb{R}^2$, 
and $\mathcal{F}_t = \{x_0, x_1, \cdots, x_t, y_0, y_1, \cdots, y_t,\pi_0,\pi_1,\cdots,\pi_{t-1}\}$.

The function $J^*(t,x_t,y_t)$ is called the optimal value function of problem (\ref{eq:Vpi})
$$
\begin{aligned}
J^*(t,x_t,y_t) 
& = \min _{\pi_t, \cdots, \pi_{T-1}} J(t,x_t,y_t)  \\
& = \min _{\pi_t, \cdots, \pi_{T-1}} \mathbb{E}\left[\begin{pmatrix}x_T\\y_T \end{pmatrix}' Q_T \begin{pmatrix}x_T \\ y_T \end{pmatrix}
+ \lambda \sum_{s=t}^{T-1} \int_{\mathbb{R}^m} \pi_s(u) \ln \pi_s(u) \mathrm{d}u \bigg|\mathcal{F}_t\right],
\end{aligned} 
$$ 
where $J(T,x_T,y_T) = \begin{pmatrix} x_T \\ y_T \end{pmatrix}' Q_T \begin{pmatrix} x_T \\ y_T\end{pmatrix}$.

Furthermore, we can derive Bellman's equation as
$$
\begin{aligned}
J^*(t,x_t,y_t) = & \min _{\pi_t(u)}\mathbb{E}\left[J^*(t+1,x_{t+1},y_{t+1}) + \lambda  \int_{\mathbb{R}^m} \pi_t(u) \ln \pi_t(u) \mathrm{d}u \bigg| \mathcal{F}_t\right].    
\end{aligned}
$$


\begin{theorem}\label{thm:general} 
At period $t$, the optimal value function can be presented by 
\begin{equation}
\begin{aligned}
J^*(t,x_t,y_t)
= & \begin{pmatrix}x_t \\ y_t\end{pmatrix}' P_t \begin{pmatrix}x_t \\ y_t\end{pmatrix}
      + \frac{\lambda}{2}\sum_{k=t}^{T-1}\ln \left[{\bigg(\frac{1}{\pi\lambda}\bigg)}^m|G_{k}|\right]
\end{aligned}	
\end{equation} 
and the corresponding optimal feedback control $u$ follows Gaussian distribution, with its density function given by 
\begin{equation}
\pi_t^*(u)=\mathcal{N}\left(-G_t^{-1}H_t' \begin{pmatrix} x_t\\y_t\end{pmatrix},~\frac{\lambda}{2}G_t^{-1}\right), 
\end{equation}
where
$$
\begin{aligned}
P_{t} & = F_t - H_t G_t^{-1}H_t', \\ 
F_{t} & = \begin{pmatrix} A & 0 \\ 0 & \bar A \end{pmatrix}' P_{t+1} \begin{pmatrix} A &0\\0&\bar A \end{pmatrix}
               + \begin{pmatrix}C & 0 \\ 0 & \bar C \end{pmatrix}' P_{t+1} \begin{pmatrix}C &0\\0&\bar C \end{pmatrix}, \\ 
H_{t} & = \begin{pmatrix} A & 0 \\ 0  & \bar A \end{pmatrix}' P_{t+1}\begin{pmatrix} B \\0\end{pmatrix} 
               + \begin{pmatrix}C & 0 \\ 0 & \rho\bar C \end{pmatrix}' P_{t+1} \begin{pmatrix}D \\ 0\end{pmatrix}, \\
G_{t} & = \begin{pmatrix} B \\ 0\end{pmatrix}' P_{t+1}\begin{pmatrix} B \\ 0\end{pmatrix}
               + \begin{pmatrix}D \\ 0\end{pmatrix}' P_{t+1} \begin{pmatrix}D \\ 0 \end{pmatrix}, 
  \end{aligned}
$$
for $t=0,1,\cdots T-1$ and $P_T = Q_T$. 
\end{theorem}

\textit{Proof}. It can be found in Appendix. 
\hfill $\Box$

\vspace{0.5cm} 


\begin{theorem}\label{thm:Q}
If $Q_T = \begin{pmatrix} 1 & -1 \\ -1 & 1\end{pmatrix}$, then $J^*(t,x_t,y_t)$ can be further presented by 
\begin{equation}
\begin{aligned}\label{eq:optimal_j}
J^*(t,x_t,y_t) 
& = \begin{pmatrix}x_t \\ y_t\end{pmatrix}' P_t \begin{pmatrix}x_t \\ y_t\end{pmatrix}
      + \frac{\lambda}{2}\ln\bigg(\frac{1}{\pi\lambda}\bigg)^m(T-t) + \frac{\lambda}{2}\ln\left|B'B + D'D\right|(T-t) \\ 
& \quad + \frac{\lambda}{4}\ln\left((A^2+C^2) - (AB+CD)(B'B + D'D)^{-1}(AB+CD)'\right)(T-t-1)(T-t) 
\end{aligned}	
\end{equation} 
and the density function of the feedback control $u$ is given by
\begin{equation}\label{eq:optimal_policy}
\pi_t^*(u)=\mathcal{N}\left(u \bigg\vert -G_t^{-1}H_t' \begin{pmatrix} x_t\\y_t\end{pmatrix},~\frac{\lambda}{2}G_t^{-1}\right), 
\end{equation}
where
$$
\begin{aligned}
P_{t} 
& = \begin{pmatrix} P_{t,11} & P_{t,12} \\ P_{t,21} & P_{t,22}\end{pmatrix}, \\ 
P_{t,11}
& = \left((A^2+C^2) - (AB+CD)(B'B + D'D)^{-1}(AB+CD)'\right)^{T-t}, \\ 
P_{t,12}
& = P_{t,21} = -\left(A\bar A+C\bar C - (\bar AB + \rho\bar C D)(B'B + D'D)^{-1}(AB+CD)'\right)^{T-t}, \\ 
P_{t,22}
& = -\Bigg(\dsp\sum_{j=t}^{T-1}\left(\frac{(A\bar A + C\bar C - (\bar AB + \rho\bar C D)(B'B + D'D)^{-1}(AB + CD)')^2}{A^2 + C^2 - (AB + CD)(B'B + D'D)^{-1}(AB + CD)' }\right)^{T-j-1} \\ 
& \quad \times[(\bar AB + \rho\bar C D)(B'B + D'D)^{-1}(\bar AB + \rho\bar C D)'](\bar A^2 + \bar C^2)^{j-t}\Bigg) + (\bar A^2 + \bar C^2)^{T-t}, \\ 
G_{t} 
& = (B'B + D'D)\left((A^2+C^2) - (AB+CD)(B'B + D'D)^{-1}(AB+CD)'\right)^{T-t-1}, \\ 
G_{t}^{-1}H_{t}' 
& = \begin{pmatrix}(B'B + D'D)^{-1}(AB+CD)' & & g_t(B'B + D'D)^{-1}(\bar AB+ \rho\bar CD)'\end{pmatrix}, \\ 
g_t 
& = -\left(\frac{A\bar A + C\bar C - (AB + CD)(B'B + D'D)^{-1}(\bar AB + \rho\bar CD)'}{(A^2+C^2) - (AB+CD)(B'B + D'D)^{-1}(AB+CD)'}\right)^{T-t-1}, 
\end{aligned}
$$ 
for $t=0,1,\cdots T-1$. 
\end{theorem}

\textit{Proof}. It can be found in Appendix. 
\hfill $\Box$

\section{Mean--Variance Model of Asset-liability Management}

In this section, we consider a simple financial application. Let us recall the classical discrete-time mean--variance (MV) problem in Li-Ng (2000). 
We consider an investment market consists of one risk-free asset, one risky asset and one liability.
At time period $t$, the given deterministic return of the risk-free asset, the excess return of the risky asset, and the random return of the liability are denoted by $r_f (>1)$, $r_t$ and $q_t$, respectively. An investor reallocates her portfolio at the beginning of each of the following $T-1$ consecutive periods. Also, assume that $r_t$ and $q_t$ are statistically independent at different time periods. 
Let $X_t$ and $l_t$ be the wealth and liability of the investor at the beginning of period $t$, respectively, then $X_t-l_t$ is the net wealth. At period $t$, if $u_t$ is the amount invested in the risky asset, then $X_t-u_t$ is the amount invested in the risk-free asset. In this project, assume that the liability is exogenous, i.e., it is uncontrollable and cannot be affected by the investor's strategy. 

The discrete-time mean--variance model of asset-liability management is to seek the best strategy, $u_t^*, t=0,1, \cdots, T-1$, which is the solution of the following dynamic wealth problem,
\begin{equation}\label{eq:mv} 
\left\{\begin{array}{rl}
\dsp\min_{u} & \mbox{Var}\left(X_T^u-l_T\right) \equiv \mathbb{E}[(X_T^u - l_T)^2] - d^2, \\ [2mm] 
\mbox{s.t.} & \mathbb{E}\left[X_T^u-l_T\right]=d, \\ [2mm] 
& X_{t+1}^u = r_fX_t^u + r_tu_t, \\ [2mm] 
& l_{t+1} = q_t l_t,  \quad\quad\quad\quad\quad\quad\quad\quad\quad\quad \mbox{for }  t=0,1, \cdots, T-1, 
\end{array}\right.
\end{equation} 
with an initial endowment $X_0^u = X_0$ and initial liability $l_0$. 

Introducing a Lagrange multiplier $2\gamma > 0$, we transform (\ref{eq:mv}) into an unconstrained problem  
\begin{equation}\label{eq:mv1} 
\min_{u} ~ \mathbb{E}[(X_T^u - l_T)^2] - d^2 -2\gamma\left(\mathbb{E}[X_T^u - l_T]-d\right) 
= \min_{u} ~ \mathbb{E}[(X_T^u - \gamma - l_T)^2] - (\gamma - d)^2. 
\end{equation} 
The above problem can be solved analytically, whose solution $u^* = \{u_t^*, 0 \leq t \leq T-1\}$ depends on $\gamma$. Then the original constraint $\mathbb{E}[X_T^{u^*} - l_T] = d$ determines the value
of $\gamma$. We refer a detailed derivation to Li et al. (2017).

Set $x_t^u := X_t^u - \gamma r_f^{-(T-t)}$. Then we transform problem (\ref{eq:mv}) into the following unconstrained control problem 
\begin{equation}\label{eq:mv2} 
\left\{\begin{array}{rl}
\dsp\min_{u} & \mathbb{E}\left[(x_T^u-l_T)^2\right], \\ [2mm] 
\mbox{s.t.} & x_{t+1}^u = r_fx_t^u + r_tu_t, \\ [2mm] 
& l_{t+1} = q_t l_t,  \quad\quad\quad\quad\quad\quad t=0,1, \cdots, T-1,
\end{array}\right.
\end{equation} 
where $x_0^u = X_0^u - \gamma r_f^{-T}$. 

Set $r_f := A$, $r_t := B + Dw_t^x$ and $q_t := \bar A + \bar Cw_t^l \in \mathbb{R}$. Then problem (\ref{eq:mv2}) becomes 
\begin{equation}\label{eq:mv3} 
\left\{\begin{aligned}
\dsp\min_{u} \quad & \mathbb{E}\left[(x_T^u-l_T)^2\right], \\ 
\mbox{s.t.} \quad & x_{t+1}^u = Ax_t^u + Bu_t + Du_tw_t^x, \\ 
& l_{t+1} = \bar Al_t + \bar Cl_tw_t^l, \quad\quad\quad\quad\quad\quad t=0,1, \cdots, T-1. 
\end{aligned}\right.
\end{equation} 

Define the value function $J(t, x_t, l_t)$ under any given policy $\pi$: 
\begin{equation}
\begin{aligned}
J(t, x_t, l_t) 
& = \mathbb{E}\left[\begin{pmatrix}x_T \\ l_T \end{pmatrix}' Q_T \begin{pmatrix}x_T \\ l_T \end{pmatrix} 
+ \lambda \sum_{t=0}^{T-1} \int_{\mathbb{R}^m} \pi_t(u) \ln \pi_t(u) \mathrm{d} u\bigg| \mathcal{F}_t \right], 
\end{aligned}
\end{equation}
where $Q_T = \begin{pmatrix} 1 & -1 \\ -1 & 1\end{pmatrix}$.


Applying Theorem \ref{thm:Q}, we get the following proposition. 

\begin{proposition}\label{prop:Q} 
At period $t$, the optimal value function can be presented by 
\begin{equation}
\begin{aligned}
J^*(t,x_t,l_t) 
& = \begin{pmatrix}x_t \\ l_t\end{pmatrix}' P_t \begin{pmatrix}x_t \\ l_t\end{pmatrix}
      + \frac{\lambda}{2}\ln\bigg(\frac{1}{\pi\lambda}\bigg)(T-t) + \frac{\lambda}{2}\ln\left(B^2 + D^2\right)(T-t) \\ 
& \quad + \frac{\lambda}{4}\ln\left(\frac{A^2D^2}{B^2+D^2}\right)(T-t-1)(T-t) \\ 
\end{aligned}	
\end{equation} 
and the density function of the feedback control $u$ is given by
\begin{equation}
\pi_t^*(u)=\mathcal{N}\left(u \bigg\vert -G_t^{-1}H_t' \begin{pmatrix} x_t\\l_t\end{pmatrix},~\frac{\lambda}{2}G_t^{-1}\right), 
\end{equation} 
where
$$
\begin{aligned}
P_{t} 
& = \begin{pmatrix} P_t^{11} & P_t^{12} \\ P_t^{21} & P_t^{22}\end{pmatrix}, \\ 
P_{t,11}
& = \left(\frac{A^2D^2}{B^2+D^2}\right)^{T-t}, \\ 
P_{t,12} 
& = P_{t,21} = -\left(\frac{A\bar AD^2-\rho AB\bar CD}{B^2+D^2}\right)^{T-t} = -A^{T-t}\left(\frac{\bar AD^2-\rho B\bar CD}{B^2+D^2}\right)^{T-t}, \\ 
\end{aligned}
$$ 
$$
\begin{aligned}
P_{t,22}
& = -\Bigg(\dsp\sum_{j=t}^{T-1}\left(\frac{(\bar AD - \rho B\bar C)^2}{B^2 + D^2}\right)^{T-j-1} 
       \frac{(\bar AB + \rho\bar C D)^2}{B^2 + D^2}(\bar A^2 + \bar C^2)^{j-t}\Bigg) + (\bar A^2 + \bar C^2)^{T-t}, \\ 
G_{t} 
& = (B^2 + D^2)\left(\frac{A^2D^2}{B^2 + D^2}\right)^{T-t-1}, \\ 
G_{t}^{-1}H_{t}' 
& = \begin{pmatrix}\dsp\frac{AB}{B^2 + D^2} & \quad & \dsp -\left(\frac{\bar AD - \rho B\bar C}{AD}\right)^{T-t-1}\frac{\bar AB+ \rho\bar CD}{B^2 + D^2}\end{pmatrix}, 
\end{aligned}
$$ 
for $t=0,1,\cdots T-1$. 
\end{proposition}

\vspace{0.5cm} 

\section{Reinforcement Learning Algorithm} 

Having studied the optimal solution of the discrete-time LQ control problem in the previous sections, we next design an RL algorithm to learn the solution directly and output optimal allocation strategies. First, we establish a policy improvement theorem and the corresponding convergence result. Second, we design a self-correcting scheme to learn the true Lagrange multiplier $\gamma$. Furthermore, it is notable that this RL algorithm needs to include estimating model parameters related to dynamics, which are challenging to calculate with precision.

\subsection{Policy improvement and policy convergence}

To prove policy improvement and policy convergence, we first introduce the following lemma, which will be used in later proof.

\begin{lemma}\label{lem:KLN}
Suppose that $\mathbf{\bar\pi} = \{\bar\pi_0, \bar\pi_1, \cdots, \bar\pi_{T-1}\}$ is an arbitrarily given admissible feedback control policy, such as 
\begin{align*}
\bar\pi_{t}(u;x_t,l_t) 
& = \mathcal{N} \left(u \bigg\vert K\begin{pmatrix}x_{t} \\ l_{t} \end{pmatrix}, \lambda LN^{T-t-1}\right), \quad 
\mbox{for } ~ t = 0, 1, \cdots, T-1. 
\end{align*}
Then 
\begin{equation*}
\begin{aligned}
J^{\bar\pi}\left(t,x_{t},l_{t}\right) 
& = \begin{pmatrix}x_{t} \\ l_{t}\end{pmatrix}' M_{t} \begin{pmatrix}x_{t} \\ l_{t}\end{pmatrix} 
+ \frac{\lambda L(B^2+D^2)\left(1-(N\bar m)^{T-t}\right)}{1-N\bar m} \\
& \quad - \frac{\lambda}{2}\ln(2\pi\lambda L)(T-t)  - \frac{\lambda}{2}(T-t) - \frac{\lambda}{2}\ln(N)\frac{(T-t-1)(T-t)}{2} \\ 
& = \begin{pmatrix}x_{t} \\ l_{t}\end{pmatrix}' M_{t} \begin{pmatrix}x_{t} \\ l_{t}\end{pmatrix} 
+ f(t), 
\end{aligned}
\end{equation*}
where 
{\small 
$$
\begin{aligned} 
M_{t} 
& = \begin{pmatrix} (A^2 + 2ABK_1 + (B^2+D^2)K_1'K_1)M_{t+1,11} 
& A\bar AM_{t+1,12} + (2ABK_2 + (B^2+D^2)K_1'K_2)M_{t+1,11} \\
(A\bar A + 2(\bar AB + \rho\bar CD)K_1)M_{t+1,21} + (B^2+D^2)K_2'K_1M_{t+1,11} & M_{t, 22} 
\end{pmatrix} \\ 
& = \begin{pmatrix} M_{t,11} & M_{t,12} \\
M_{t,21} & M_{t,22}
\end{pmatrix} 
= \begin{pmatrix} \bar m^{T-t} & M_{t,12} \\
M_{t,21} & M_{t,22}
\end{pmatrix}, 
\end{aligned} 
$$ 
} 
$$
M_{t,22} = (\bar A^2 + \bar C^2)M_{t+1,22}  + 2(\bar AB + \rho\bar CD)K_2M_{t+1,21} + (B^2+D^2)K_2'K_2M_{t+1,11} 
$$ 
and 
$$
\begin{aligned}
f(t) 
& = \frac{\lambda L(B^2+D^2)\left(1-(N\bar m)^{T-t}\right)}{1-N\bar m} 
- \frac{\lambda}{2}\ln(2\pi\lambda L)(T-t)  - \frac{\lambda}{2}(T-t) - \frac{\lambda}{2}\ln(N)\frac{(T-t-1)(T-t)}{2}. 
\end{aligned}
$$
\end{lemma}

\textit{Proof}. It can be found in Appendix. 
\hfill $\Box$

\vspace{0.5cm} 

We design the policy improvement scheme to adjust the existing policy to enhance the value function. Next, we establish a policy improvement theorem that guarantees the iterated value function to be non-increasing for our minimization problem. Then, we prove that the iterated value function eventually converges to the optimal value function. The following results of a policy improvement theorem and policy convergence deal with a discrete-time exploratory asset-liability management problem. 


\begin{theorem}\label{thm:update} 
Let $\pi_{t}^0(u;x_t,l_t) = \bar\pi_{t}(u;x_t,l_t)$. 
Then, using $\pi_t^{j+1}(u; x_t, l_t) = \dsp\arg\min_{\pi_t^j(u)} J^{\pi^j}(t, x_t, l_t)$ to update the feedback policy and making this iteration for $j$ times, we can get 
$$
\pi_t^j(u; x_t, l_t)=\mathcal{N}\left(u \bigg\vert -G_t^{-1}H_t' \begin{pmatrix} x_t\\l_t\end{pmatrix},~\frac{\lambda}{2(B^2 + D^2)(\bar m)^{T-(t+j)}}\left(\frac{B^2 + D^2}{A^2D^2}\right)^{j-1}\right)
$$ 
and the corresponding value function 
$$
\begin{aligned}
J^{\pi^j}(t,x_{t},l_{t})
& = \begin{pmatrix} x_{t} \\ l_{t} \end{pmatrix}' M_{t+j} \begin{pmatrix} x_{t} \\ l_{t} \end{pmatrix} 
+ \frac{\lambda}{2}j\ln\bigg(\frac{B^2+D^2}{\pi\lambda}\bigg) + \frac{\lambda}{4}(j-1)j\ln\left(\frac{A^2D^2}{B^2+D^2}\right) \\ 
& \quad + \frac{\lambda}{2}j\ln(\bar m)(T-t-j) + f(t+j), 
\end{aligned}
$$ 
where
$$
\begin{aligned} 
G_{t}^{-1}H_{t}' 
& = \begin{pmatrix}\dsp\frac{AB}{B^2 + D^2} & \quad & \dsp -\left(\frac{\bar AD - \rho B\bar C}{AD}\right)^{T-t-1}\frac{\bar AB+ \rho\bar CD}{B^2 + D^2}\frac{M_{t+j,21}}{\bar m^{T-(t+j)}}\end{pmatrix}, 
\end{aligned}
$$ 
for $t=0,1,\cdots T-1$. 
\end{theorem}

\textit{Proof}. It can be found in Appendix. 
\hfill $\Box$

\subsection{Algorithm Design}	
The algorithm contains three parts: \textit{Initialization, policy evaluation and policy improvement.}
\paragraph{Initialization}
We first designate $ J^{\theta}$ and $\pi^{\theta}$ to represent the parameterized objective function and policy, respectively, where $\theta$ are parameters to be learned. Define $ J^\theta$ as follows
\begin{equation}\label{eq:j_theta}
\begin{aligned}
J^{\theta}(t,x_t,y_t) 
& = \theta_1^{T-t} x_t^2 -2\theta_2^{T-t} x_tl_t
+\left(\dsp\sum_{j=t}^{T-1}\left(\frac{\theta_2^2}{\theta_1}\right)^{T-j-1}\theta_3^2\theta_4\theta_5^{j-t}+\theta_5^{T-t}\right)l_t^2 \\ 
& -\frac{\lambda}{2}\ln\theta_4(T-t) + \frac{\lambda}{4}\ln\theta_1 (T-t-1)(T-t)  +\frac{\lambda}{2}\ln\bigg(\frac{1}{\pi\lambda}\bigg)(T-t)
\end{aligned}
\end{equation}
and $\pi_t^*(u)$ as follows
\begin{equation}\label{eq:pi_theta}
\pi_t^{\theta}(u)=\mathcal{N}\left(\sqrt{(A^2-\theta_1)\theta_4}x_t -\left(\frac{\theta_1 }{\theta_2}\right)^{T-t-1}\theta_3\theta_4y_t, ~~ \frac{\lambda}{2}\theta_4\theta_1^{1-t-T} \right). 
\end{equation}
Then we need to learn $\theta_1,\theta_2,\theta_3,\theta_4,\theta_5$.

\paragraph{Policy Evaluation and Policy Improvement} 

By Bellman equation, we have the following 
\begin{equation}\label{eq:Bellman}
	J^{\pi}(t,x_t,y_t) = \mathbb{E}\bigg[J^{\pi}(t+1,x_{t+1},y_{t+1}) +\lambda  \int_{\mathbb{R}} \pi_t(u) \ln \pi_t(u) \mathrm{d} u \bigg| \mathcal{F}_t \bigg].
\end{equation}
Re-arranging equation (\ref{eq:Bellman}), we obtain
\begin{equation*}
	\mathbb{E}\bigg[J^{\pi}(t+1,x_{t+1},y_{t+1})-J^{\pi}(t,x_t,y_t) +\lambda  \int_{\mathbb{R}} \pi_t(u) \ln \pi_t(u) \mathrm{d} u \bigg| \mathcal{F}_t \bigg]= 0.
\end{equation*}
Define the Bellman's error as
\begin{equation*}
	\delta_t = J^{\pi}(t+1,x_{t+1},y_{t+1})-J^{\pi}(t,x_t,y_t) +\lambda  \int_{\mathbb{R}} \pi_t(u) \ln \pi_t(u) \mathrm{d}u.
\end{equation*}
The objective of policy evaluation is to minimize the summation of Bellman's error $\delta_t$. We then minimize
\begin{equation*}
	\begin{aligned}
		L(\theta) &= \frac{1}{2} \mathbb{E}\left[ \sum_{t=0}^{T-1}\delta_t^2 \right]
		= \frac{1}{2} \mathbb{E}\left[ \sum_{t=0}^{T-1} \bigg(J^{\theta}_{t+1} -J^{\theta}_{t} +\lambda  \int_{\mathbb{R}} \pi_t(u) \ln \pi_t(u) \mathrm{d} u\bigg)^2\right]\\
	\end{aligned}
\end{equation*}
We adopt stochastic gradient descent (SGD) to minimize $L(\theta)$. Then we have
\begin{align}
	&\frac{\partial L}{\partial \theta_1} = \mathbb{E}\left[ \sum_{t=0}^{T-1}\delta_t \left(\frac{\partial P_{t+1,11}}{\partial\theta_1}x^2_{t+1}+\frac{\partial P_{t+1,22}}{\partial\theta_1}l^2_{t+1}-\frac{\partial P_{t,11}}{\partial\theta_1}x^2_{t}-\frac{\partial P_{t,22}}{\partial\theta_1}l^2_{t}\right) \right], \label{eq:dtheta1} \\ 
	&\frac{\partial L}{\partial \theta_2} = \mathbb{E}\left[ \sum_{t=0}^{T-1}\delta_t \left(2\frac{\partial P_{t+1,12}}{\partial\theta_2}x_{t+1}l_{t+1}+\frac{\partial P_{t+1,22}}{\partial\theta_2}l^2_{t+1} -2\frac{\partial P_{t,12}}{\partial\theta_2}x_{t}l_{t}-\frac{\partial P_{t,22}^{}}{\partial\theta_2}l^2_{t}\right)\right], \label{eq:dtheta2} \\ 
	&\frac{\partial L}{\partial \theta_3} = \mathbb{E}\left[ \sum_{t=0}^{T-1}\delta_t \left(\frac{\partial P_{t+1,22}^{}}{\partial\theta_3}x_{t+1}^2-\frac{\partial P_{t,22}^{}}{\partial\theta_3}x_{t}^2\right) \right], \label{eq:dtheta3} \\ 
&\frac{\partial L}{\partial \theta_4}= \mathbb{E}\left[ \sum_{t=0}^{T-1}\delta_t \left(\frac{\partial P_{t+1,22}^{}}{\partial\theta_4}x_{t+1}^2-\frac{\partial P_{t,22}^{}}{\partial\theta_4}x_{t}^2\right) \right], \label{eq:dtheta4} \\
	&\frac{\partial L}{\partial \theta_5} = \mathbb{E}\left[ \sum_{t=0}^{T-1}\delta_t \left(\frac{\partial P_{t+1,22}^{}}{\partial\theta_5}x_{t+1}^2-\frac{\partial P_{t,22}^{}}{\partial\theta_5}x_{t}^2\right) \right], \label{eq:dtheta5} 
\end{align}
where
\begin{equation*}
\begin{aligned}
\frac{\partial P_{t,11}}{\partial\theta_1} 
&= (T-t)\theta_1^{T-t-1}, \\
\frac{\partial P_{t,22}}{\partial\theta_1}
&=-\dsp\sum_{j=t}^{T-1}(T-j-1)\theta_1^{-1}\left(\frac{\theta_2^2}{\theta_1}\right)^{T-j-1}\theta_3^2\theta_4\theta_5^{j-t}, \\
\frac{\partial P_{t,12}}{\partial\theta_2}
&= (T-t)\theta_2^{T-t-1}, \\
\frac{\partial P_{t,22}}{\partial\theta_2}
&= \dsp\sum_{j=t}^{T-1}(T-j-1)2\theta_2^{-1}\left(\frac{\theta_2^2}{\theta_1}\right)^{T-j-1}\theta_3^2\theta_4\theta_5^{j-t}, \\
\frac{\partial P_{t,22}}{\partial\theta_3}
&= \dsp\sum_{j=t}^{T-1}\left(\frac{\theta_2^2}{\theta_1}\right)^{T-j-1}2\theta_3\theta_4\theta_5^{j-t}, \\
\frac{\partial P_{t,22}}{\partial\theta_4}
&=\dsp\sum_{j=t}^{T-1}\left(\frac{\theta_2^2}{\theta_1}\right)^{T-j-1}\theta_3\theta_5^{j-t}, \\
\frac{\partial P_{t,22}}{\partial\theta_5}
&=\dsp\sum_{j=t}^{T-1}(j-t)\left(\frac{\theta_2^2}{\theta_1}\right)^{T-j-1}\theta_3^2\theta_4\theta_5^{j-t-1}+(T-t)\theta_5^{T-t-1}.
\end{aligned}
\end{equation*}

\paragraph{Lagrange Multiplier} Finally, we provide a scheme for learning the Lagrange multiplier $\gamma$ using the learning rate $\eta_\gamma$. Then, we have
$
\gamma_{n+1} =  \gamma_{n}- \eta_\gamma (x_T + \gamma_n -l_T-b).
$
In implementation, we replace $x_T$ by the sample average 
$\frac{1}{\widetilde N} \dsp\sum_{j=1}^{\widetilde N}x^j_T$, where $\widetilde N$ is the sample
size and $x_T$ are the most recent $\widetilde N$ terminal wealth values obtained at the time when
$\gamma$ is to be updated.

\vspace{1cm} 

\begin{algorithm}[H]
	\caption{Reinforcement Learning Algorithm for Discrete-Time LQ}\label{alg:cap2}
	\KwIn{Initial value of parameters $\theta$, learning rate $\eta$ for $\theta$ and  $\eta_\gamma$ for $\gamma$, investment horizon $T$, time step $\Delta T$, initial state $(x_0,y_0)$,  number of training episodes $\widetilde M$, sample average size $\widetilde N$}
	\For{$k=1$ to $\widetilde M$}{
		\For{$t=1$ to  $T$ (with time step $\Delta T) $}{
			Sample$(t,x_t,l_t)~$ from real market \\
		}
		\For{$i=1$ to $5$}{
			Update $\theta_i\leftarrow \theta_i - \eta \frac{\partial L}{\partial\theta_i}$ using (\ref{eq:dtheta1})-(\ref{eq:dtheta5})\\             
		}
		\If{$k~mod ~ \widetilde N == 0 $}{
			Update $\gamma\leftarrow \gamma - \eta_\gamma \frac{1}{\widetilde N} \dsp\sum_{j=k-\widetilde N+1}^k x^j_T +\gamma -l^j_T - d$
		}
		Update $\pi_t^\theta$ using(\ref{eq:pi_theta}) \\
		Update $J_t^\theta$ using (\ref{eq:j_theta})
	}
\end{algorithm}

\section{Numerical Result}

This section simulates our RL algorithm in the discrete-time mean--variance model of asset-liability management. We first specify several parameter values. For the market mode, the annual return of risk-free asset is $1.05$. We set one risky asset's annual return and volatility as $1.4$ and $0.2$. Also, we set each period's increment and liability volatility as $1.1$ and $0.1$. We consider starting from a normalized initial wealth and initial liability as $1.0$ and $0.1$, respectively, which means we have a $10\%$ liability compared to our asset. The correlation coefficient is 0.2. Therefore, we have the following parameters 
$$
X_0 = 1, ~  l_0 = 0.1,  ~ A = 1.05,  ~ B = 0.25, ~ C = 0.2,  ~ \bar{A} = 1.1,  ~ \bar{B} = 0.1, ~ \rho = 0.2. 
$$
For this reinforcement learning algorithm, we take the hyperparameter as $\eta = 1 \times10^{-20} $. For Lagrange multiplier $\gamma$, we set  $\eta_\gamma = 5 \times10^{-2}$.

\subsection{Monthly Rebalancing}
In this subsection we adjust our portfolio monthly in 1 year and 5 years investment horizon, which means $\Delta T = \frac{1}{12}$, $T = 1$ or 5.
For $T =1 $, we set the expected return as 1.4, and for $T = 5$,  we take the expected return as 8. See figure 1.
\begin{figure}[H]
\centering
	\includegraphics[scale = 0.25]{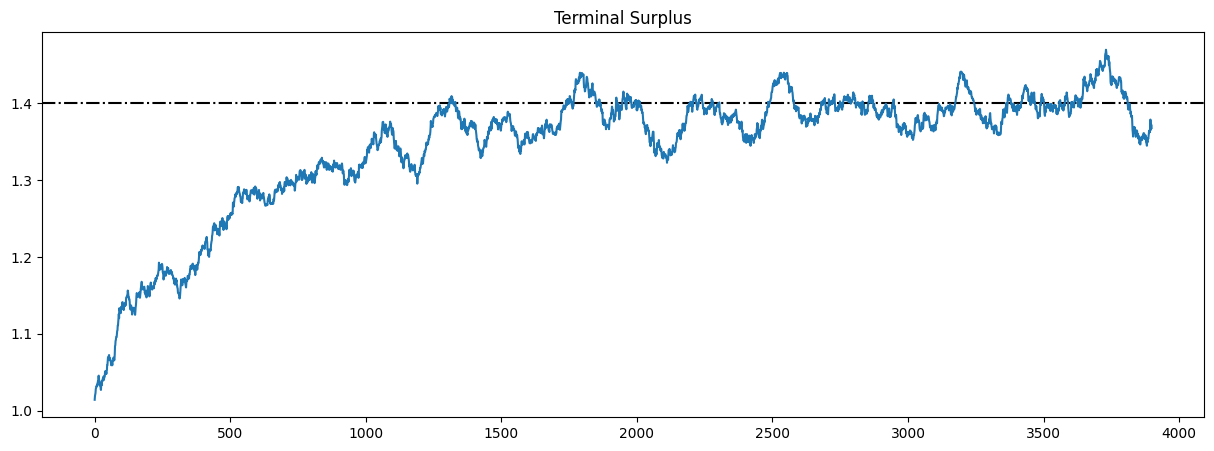}
	\includegraphics[scale = 0.25]{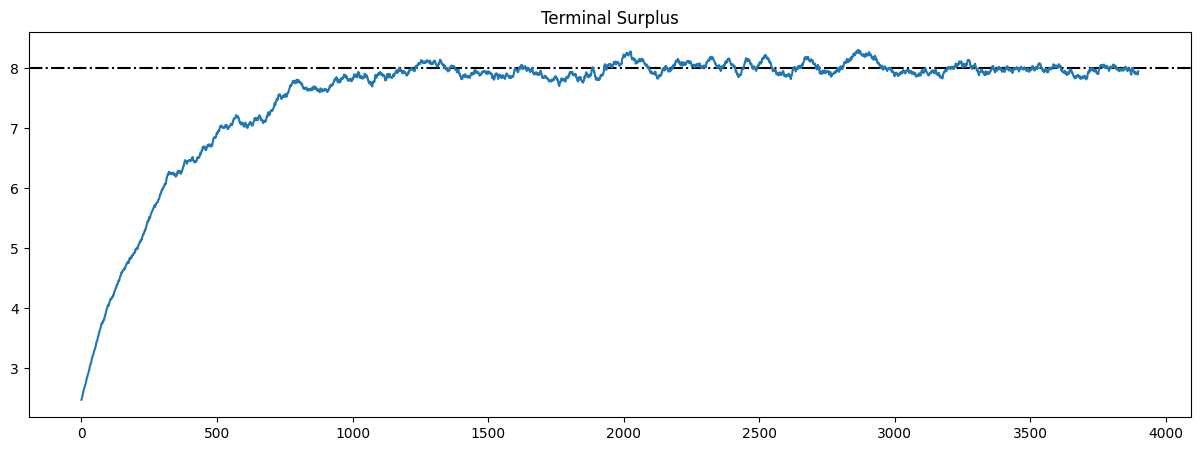}
	
	\includegraphics[scale = 0.25]{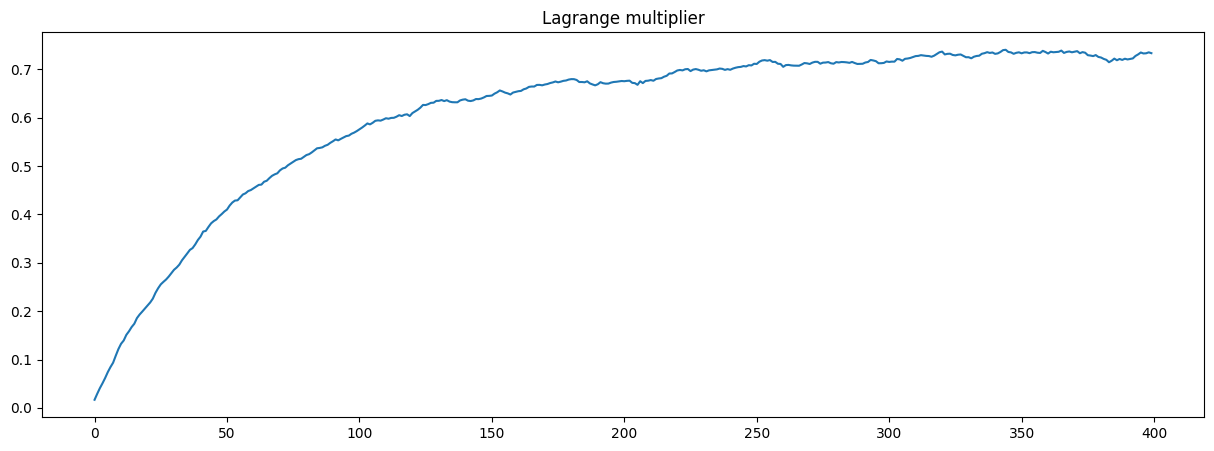}
	\includegraphics[scale = 0.25]{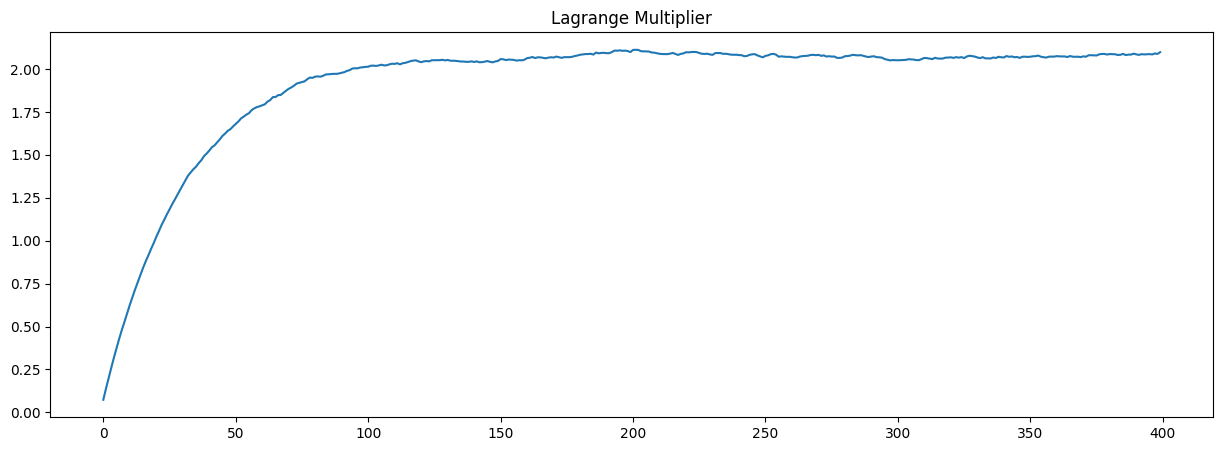}
	
	\includegraphics[scale = 0.25]{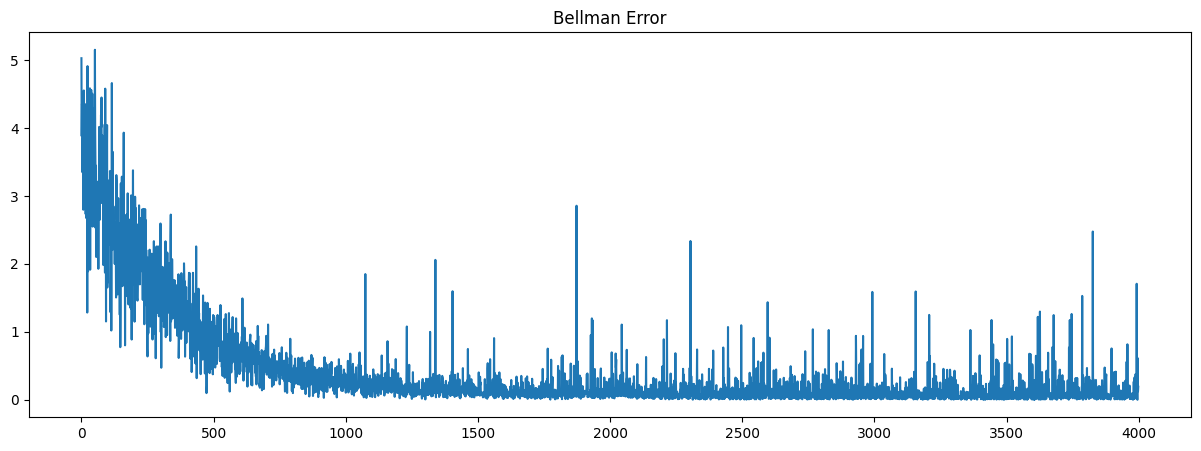}
	\includegraphics[scale = 0.25]{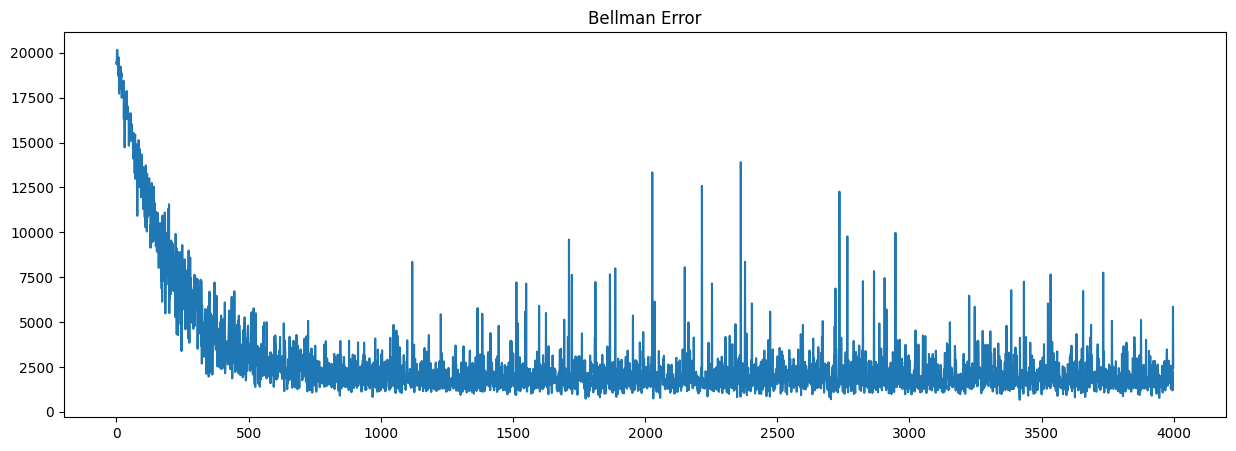}
	\label{plot:monthly}
	\caption{\small Chang of Terminal Surplus, Lagrange Multiplier and  Bellman Error in a monthly rebalancing\\(left: 1 year, right: 5years)}
\end{figure}

\subsection{Daily Rebalancing}

In this subsection, we adjust our portfolio daily in 0.5 year and 1 year  investment horizon, which means $\Delta T = \frac{1}{252}$, $T = \frac{1}{2}$ or $1$.
For $T = \frac{1}{2} $, we set the expected return as $d = 1.2$, and for $T = 1$, we take the expected return as $d=1.4$. See figure 2.

\begin{figure}[H]\centering
	\includegraphics[scale = 0.25]{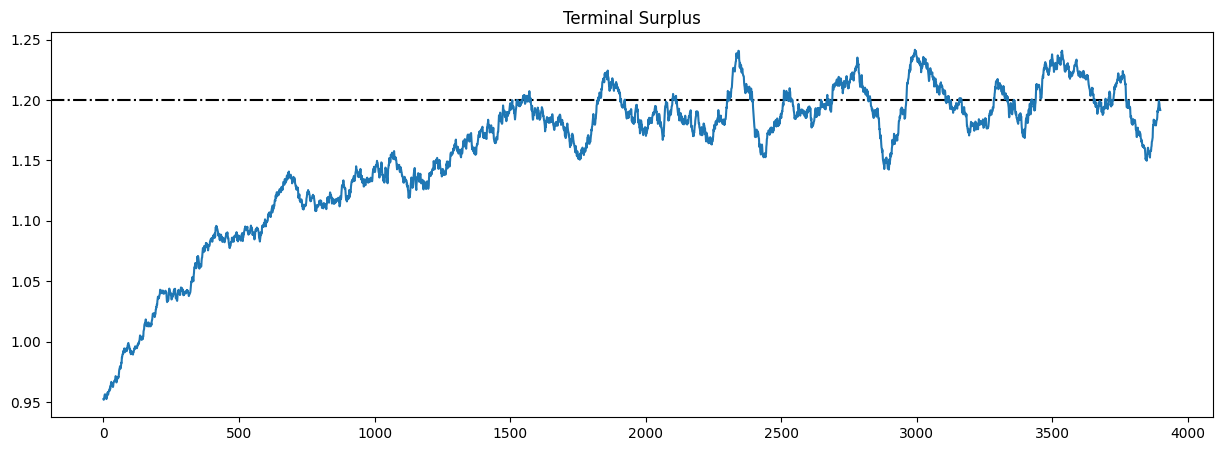}
	\includegraphics[scale = 0.25]{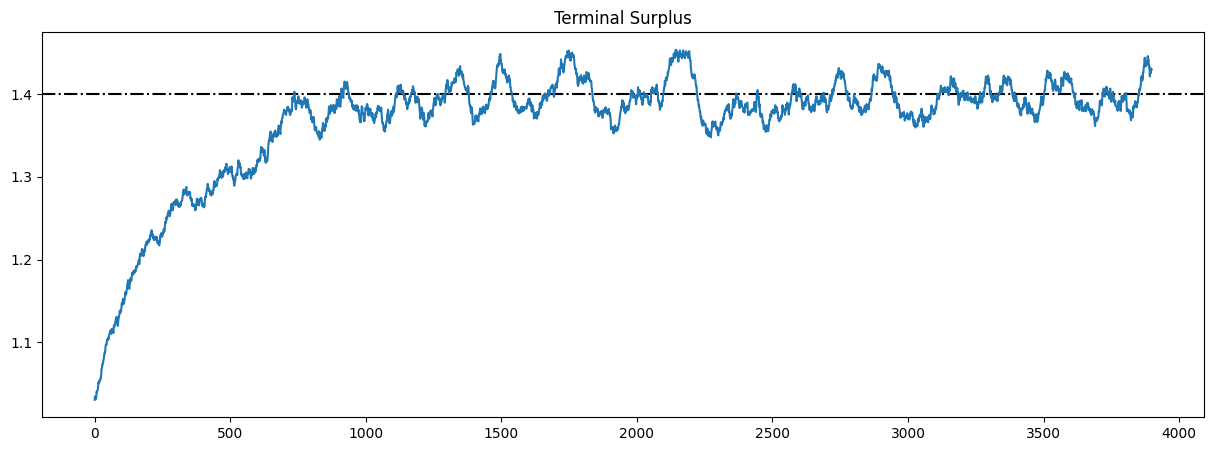}
	
	\includegraphics[scale = 0.25]{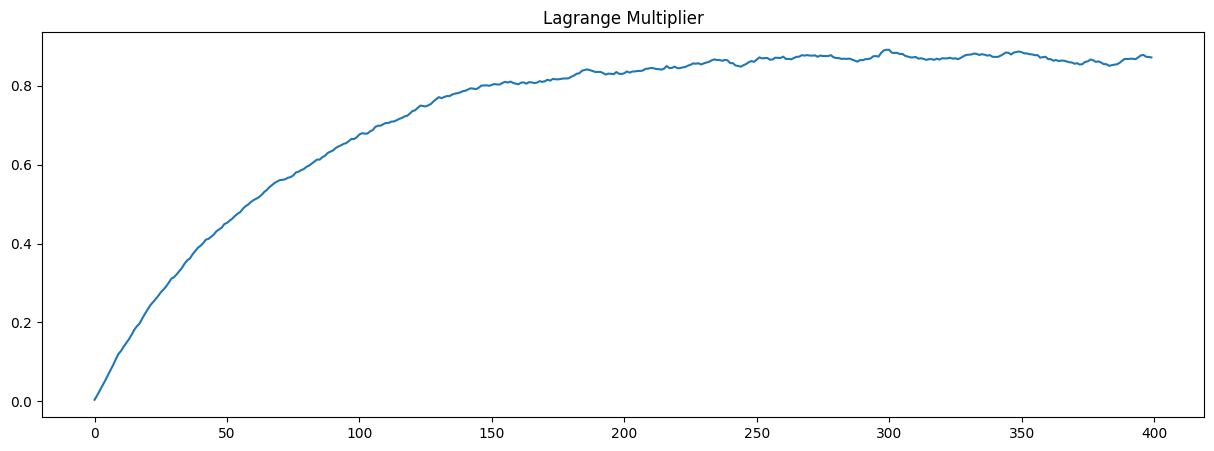}
	\includegraphics[scale = 0.25]{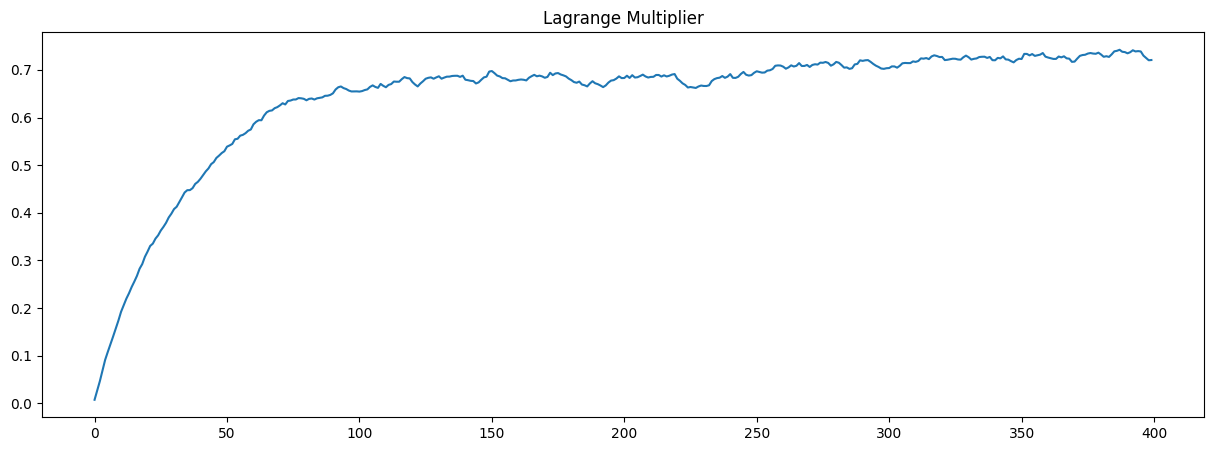}
	
	\includegraphics[scale = 0.25]{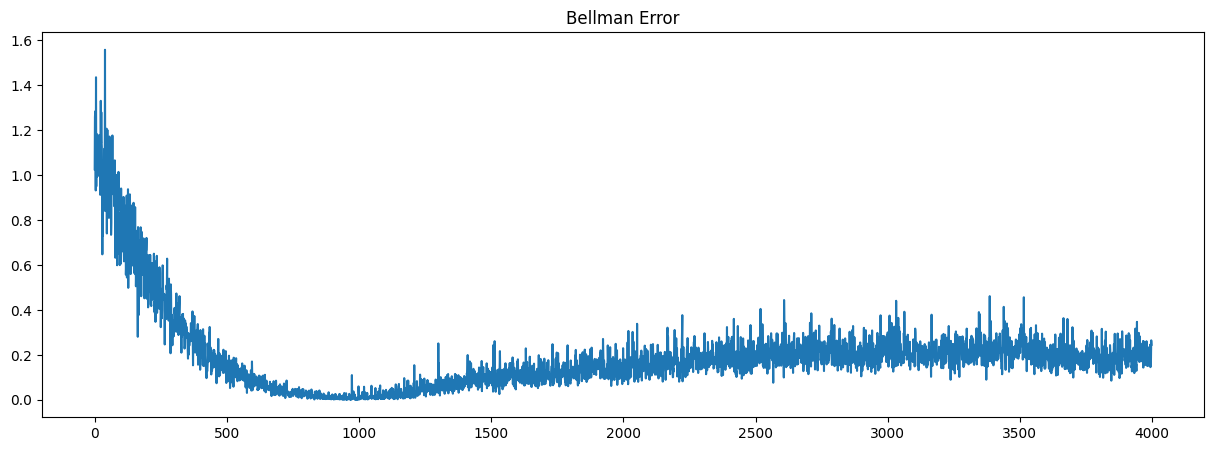}
	\includegraphics[scale = 0.25]{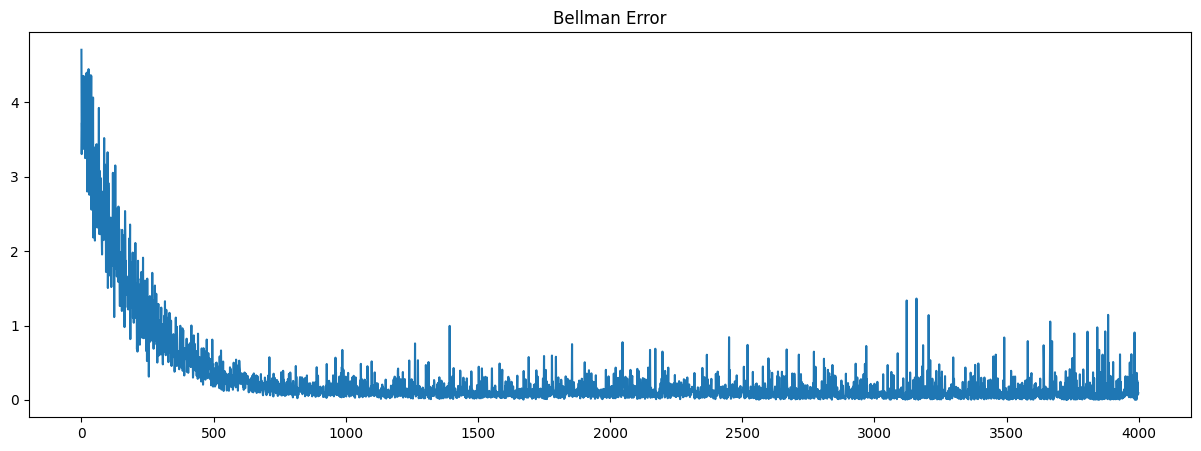}
	\label{plot:dailly}
	\caption{\small Chang of Terminal Surplus, Lagrange Multiplier and  Bellman Error in a daily rebalancing \\ (left: half year, right: 1 year)}	
\end{figure}
	
\begin{table}[H]
\begin{center}
	\begin{tabular}{|c| c| c| c|} 
	\hline
	& Sample Mean & Sample Variance & Sharp Ratio \\	
	\hline
	Monthly Rebalancing, 5 years, $d = 8$ & 8.07335 & 1.47720 & 6.60138 \\ 
	\hline
	Monthly Rebalancing, 1 year, $d = 1.4$ & 1.40067 & 0.06056 & 5.48819 \\
	\hline
	Daily Rebalancing, half year, $d = 1.2$& 1.19658 & 0.03186 & 6.42324 \\
	\hline
	Daily Rebalancing, 1 year, $d = 1.4$ & 1.39368 & 0.04920 & 6.05735 \\
	\hline
\end{tabular}
\caption{\small Sample Mean, Sample Variance and Sample Sharp Ratio of terminal wealth in different investing \\ horizon and expected return}
\end{center}
\end{table}

Our numerical result shows that our reinforcement algorithm works well in market settings and rebalancing strategies. The expected return of the sample converges with our expected return of about $2000$ in training episodes. At the same time, Lagrange multiplier $\gamma$ converges to its theoretical optimal value, and the training error converges to near $0$.


\section{Conclusion}

We study the discrete-time linear-quadratic (LQ) control model using reinforcement learning (RL). We apply the results of the discrete-time LQ model to solve the discrete-time mean-variance asset-liability management problem and prove our RL algorithm's policy improvement and convergence. However, most real-world systems are nonlinear. To overcome this challenge, we will further study a control-theoretic analysis using dynamic optimization theory techniques, extending beyond the current work. In future research, we will investigate a policy iteration RL method for the discrete-time nonlinear system to tackle real-world problems. 

\appendix

\small 

\section{Lemma}

\begin{lemma}\label{lem:P} 
Assume that $u=\{u_t, 0 \leq t<T\}$ is the control sequence and $\pi=\{\pi_t, 0 \leq t<T\}$ is the density function of $u$. Then at period $t$, we have 
\begin{equation*}
\begin{aligned}
\mathbb{E}\left[\begin{pmatrix}
        x_{t+1} \\
        y_{t+1} \\
\end{pmatrix}'
P
\begin{pmatrix}
        x_{t+1} \\
        y_{t+1} \\
\end{pmatrix} \Bigg| \mathcal{F}_{t}\right] 
& = \begin{pmatrix}
        x_{t} \\
        y_{t}
\end{pmatrix}'
\left[\begin{pmatrix} A & 0 \\ 0 & \bar A\end{pmatrix}'
P
\begin{pmatrix} A & 0 \\ 0 & \bar A\end{pmatrix}
+ \begin{pmatrix}C & 0 \\ 0 & \bar C\end{pmatrix}'
P
\begin{pmatrix}C & 0 \\ 0 & \bar C\end{pmatrix}\right]\begin{pmatrix}
        x_{t} \\
        y_{t}
\end{pmatrix} \\
& \quad + 2\dsp\int_{\mathbb{R}^m}\begin{pmatrix}
        x_{t} \\
        y_{t}
\end{pmatrix}'
\left[\begin{pmatrix}
        A & 0 \\
        0 & \bar A
\end{pmatrix}'
P
\begin{pmatrix}B \\ 0\end{pmatrix}
+ \begin{pmatrix}C & 0 \\ 0 & \rho\bar C\end{pmatrix}' 
P 
\begin{pmatrix} D \\ 0 \end{pmatrix}\right] 
u \pi_t(u) \mathrm{d}u \\
& \quad + \dsp\int_{\mathbb{R}^m} u'\left[\begin{pmatrix}B \\ 0\end{pmatrix}'
P
\begin{pmatrix}B \\ 0\end{pmatrix}
+
\begin{pmatrix} D \\ 0 \end{pmatrix}'
P
\begin{pmatrix} D \\ 0 \end{pmatrix}\right]
u \pi_t(u) \mathrm{d} u, 
\end{aligned}
\end{equation*}
where $P \in \mathbb{R}^{2\times 2}$.  
\end{lemma}

\noindent 
\textit{Proof}. 
Using the dynamic system (\ref{eq:state}), we have 
\begin{equation*}
\begin{aligned}
\mathbb{E}\left[\begin{pmatrix}
        x_{t+1} \\
        y_{t+1} \\
\end{pmatrix}'
P
\begin{pmatrix}
        x_{t+1} \\
        y_{t+1}\\
\end{pmatrix} \Bigg| \mathcal{F}_{t}\right] 
& =  \mathbb{E}\left[\begin{pmatrix}
 A x_t + B u_t + (C x_t + D u_t)w_t^x \\
\bar A y_t + \bar C y_t w_t^y
\end{pmatrix}' 
P
\begin{pmatrix}
A x_t + B u_t + (C x_t + D u_t)w_t^x \\
\bar A y_t + \bar C y_t w_t^y
\end{pmatrix}\Bigg| \mathcal{F}_{t}\right] \\
& = \mathbb{E}\left[ \begin{pmatrix}
\begin{pmatrix}A&0\\0&\bar A\end{pmatrix} \begin{pmatrix}
     x_{t}\\y_{t}\end{pmatrix}
     +\begin{pmatrix}B\\0 \end{pmatrix} u_{t}
     +\begin{pmatrix}C&0\\0&\bar C
\end{pmatrix}\begin{pmatrix}x_{t}w_t^x \\ y_{t}w_t^y 
\end{pmatrix}+\begin{pmatrix}
         D\\0
\end{pmatrix} 
u_{t}w_t^x \end{pmatrix}' \right. \\
&  \quad \left. P  \begin{pmatrix}\begin{pmatrix}A&0\\0&\bar A\end{pmatrix} \begin{pmatrix}
     x_{t}\\y_{t}\end{pmatrix}
     +\begin{pmatrix}B\\0 \end{pmatrix} u_{t}
     +\begin{pmatrix}C&0\\0&\bar C
     \end{pmatrix}\begin{pmatrix}x_{t}w_t^x \\ y_{t}w_t^y 
     \end{pmatrix}+ \begin{pmatrix}
         D\\0
     \end{pmatrix} 
         u_{t}w_t^x\end{pmatrix} \Bigg| \mathcal{F}_{t}\right] \\    
& = \begin{pmatrix}
        x_{t}\\
        y_{t}
\end{pmatrix}'
\left[\begin{pmatrix}
        A&0\\
        0&\bar A
\end{pmatrix}'
P
\begin{pmatrix}
        A & 0 \\
        0 & \bar A
\end{pmatrix}
+ \begin{pmatrix}C & 0 \\ 0 & \bar C\end{pmatrix}'
P
\begin{pmatrix}C & 0 \\ 0 & \bar C\end{pmatrix}\right]\begin{pmatrix}
        x_{t}\\
        y_{t}
\end{pmatrix} \\
& \quad + 2\dsp\int_{\mathbb{R}^m}\begin{pmatrix}
        x_{t} \\
        y_{t}
\end{pmatrix}'
\left[\begin{pmatrix}
        A & 0 \\
        0 & \bar A
\end{pmatrix}'
P
\begin{pmatrix}B \\ 0\end{pmatrix}
+ \begin{pmatrix}C & 0 \\ 0 & \rho\bar C\end{pmatrix}' 
P 
\begin{pmatrix} D \\ 0 \end{pmatrix}\right] 
u \pi_t(u) \mathrm{d}u \\
& \quad + \dsp\int_{\mathbb{R}^m} u'\left[\begin{pmatrix}B \\ 0\end{pmatrix}'
P
\begin{pmatrix}B \\ 0\end{pmatrix}
+
\begin{pmatrix} D \\ 0 \end{pmatrix}'
P
\begin{pmatrix} D \\ 0 \end{pmatrix}\right]
u \pi_t(u) \mathrm{d}u.
\end{aligned}
\end{equation*}
This completes the proof. 
\hfill $\Box$

\vspace{0.5cm} 

\section{Proof of Theorem 1}
First, using Lemma \ref{lem:P} with $t=T-1$ and $P = Q_T$, we get 
\begin{equation*}
\begin{aligned}
\mathbb{E}\left[\begin{pmatrix}
x_T \\
y_T \\
\end{pmatrix}'
Q_T
\begin{pmatrix}
x_T \\
y_T \\
\end{pmatrix} \Bigg| \mathcal{F}_{T-1}\right] 
& = \begin{pmatrix}
x_{T-1} \\
y_{T-1}
\end{pmatrix}'
\left[\begin{pmatrix}A & 0 \\ 0 & \bar A\end{pmatrix}' 
Q_T
\begin{pmatrix}A & 0 \\ 0 & \bar A \end{pmatrix}
+ \begin{pmatrix}C & 0 \\ 0 & \bar C \end{pmatrix}'
Q_T
\begin{pmatrix}C & 0 \\ 0 & \bar C\end{pmatrix}\right]\begin{pmatrix}
x_{T-1} \\
y_{T-1}
\end{pmatrix} \\
& \quad + 2\dsp\int_{\mathbb{R}^m}\begin{pmatrix}
x_{T-1} \\
y_{T-1}
\end{pmatrix}'\left[\begin{pmatrix}
A & 0 \\ 0 & \bar A
\end{pmatrix}'
Q_T
\begin{pmatrix}B \\ 0\end{pmatrix}
+ \begin{pmatrix}C & 0 \\0 & \rho\bar C\end{pmatrix}' 
Q_T \begin{pmatrix} D \\ 0 \end{pmatrix}\right] 
u \pi_{T-1}(u) \mathrm{d}u \\
& \quad + \dsp\int_{\mathbb{R}^m} u'\left[
\begin{pmatrix}B \\ 0\end{pmatrix}'
Q_T
\begin{pmatrix}B \\ 0\end{pmatrix}
+ \begin{pmatrix} D\\ 0 \end{pmatrix}'
Q_T
\begin{pmatrix} D\\ 0 \end{pmatrix}\right]
u \pi_{T-1}(u) \mathrm{d}u.
\end{aligned}
\end{equation*}

If we take $P_T = Q_T$, then we get 
$$
\begin{aligned}
& J^\pi(T-1,x_{T-1},y_{T-1}) \\
& = \mathbb{E}\left[\begin{pmatrix} x_{T} \\ y_{T} \end{pmatrix}' P_T \begin{pmatrix} x_{T} \\ y_{T} \end{pmatrix}  
      + \lambda \int_{\mathbb{R}^m} \pi_{T-1}(u) \ln \pi_{T-1}(u) \mathrm{d} u \bigg| \mathcal{F}_{T-1}\right] \\
& = \begin{pmatrix}
x_{T-1} \\
y_{T-1}
\end{pmatrix}'
\left[\begin{pmatrix}A & 0 \\ 0 & \bar A\end{pmatrix}'
P_T
\begin{pmatrix}A & 0 \\ 0 & \bar A \end{pmatrix}
+ \begin{pmatrix}C & 0 \\ 0 & \bar C \end{pmatrix}'
P_T
\begin{pmatrix}C & 0 \\ 0 & \bar C\end{pmatrix}\right]\begin{pmatrix}
x_{T-1} \\
y_{T-1}
\end{pmatrix} \\
\end{aligned}
$$
$$
\begin{aligned}
& \quad + 2\dsp\int_{\mathbb{R}^m}\begin{pmatrix}
x_{T-1} \\
y_{T-1}
\end{pmatrix}'\left[\begin{pmatrix}
A & 0 \\ 0 & \bar A\end{pmatrix}'
P_T
\begin{pmatrix}B \\ 0\end{pmatrix}
+ \begin{pmatrix}C & 0 \\0 & \rho\bar C\end{pmatrix}' 
P_T \begin{pmatrix} D \\ 0 \end{pmatrix}\right] 
u \pi_{T-1}(u) \mathrm{d}u \\
& \quad + \dsp\int_{\mathbb{R}^m} u'\left[\begin{pmatrix}B \\ 0\end{pmatrix}'
P_T
\begin{pmatrix}B \\ 0\end{pmatrix}
+ \begin{pmatrix} D\\ 0 \end{pmatrix}'
P_T
\begin{pmatrix} D\\ 0 \end{pmatrix}\right]
u \pi_{T-1}(u) \mathrm{d}u 
+ \lambda \int_{\mathbb{R}^m} \pi_{T-1}(u) \ln \pi_{T-1}(u) \mathrm{d}u \\
& = \begin{pmatrix} x_{T-1} \\ y_{T-1} \end{pmatrix} F_{T-1} \begin{pmatrix} x_{T-1} \\ y_{T-1} \end{pmatrix} 
+ \int_{\mathbb{R}^m} \left[2 \begin{pmatrix} x_{T-1} \\ y_{T-1} \end{pmatrix}' H_{T-1} u +u' G_{T-1}u
   +\lambda\ln \pi_{T-1}(u)\right] \pi_{T-1}(u) \mathrm{d}u,
\end{aligned}
$$
where
$$
\begin{aligned}
F_{T-1} & = \begin{pmatrix}A & 0 \\ 0 & \bar A\end{pmatrix}'P_T\begin{pmatrix}A & 0 \\ 0 & \bar A \end{pmatrix}
                   + \begin{pmatrix}C & 0 \\ 0 & \bar C \end{pmatrix}'P_T\begin{pmatrix}C & 0 \\ 0 & \bar C\end{pmatrix}, \\
H_{T-1} & = \begin{pmatrix}A & 0 \\ 0 & \bar A\end{pmatrix}'P_T\begin{pmatrix}B \\ 0\end{pmatrix}
                    + \begin{pmatrix}C & 0 \\0 & \rho\bar C\end{pmatrix}' P_T \begin{pmatrix} D \\ 0 \end{pmatrix}, \\
G_{T-1} & = \begin{pmatrix}B \\ 0\end{pmatrix}'P_T\begin{pmatrix}B \\ 0\end{pmatrix}
                    + \begin{pmatrix} D\\ 0 \end{pmatrix}'P_T\begin{pmatrix} D\\ 0 \end{pmatrix}. 
\end{aligned}
$$
It follows from $\frac{\partial J^{\pi}(T-1,x_{T-1},y_{T-1})}{\partial \pi_{T-1}(u)} = 0$ that we have 
$$
2 \begin{pmatrix} x_{T-1} \\ y_{T-1} \end{pmatrix}' H_{T-1} u 
+ u' G_{T-1} u + \lambda\ln \pi_{T-1}(u) + \lambda = 0.
$$
Since $\pi_{T-1}(u)$ is a probability density function of $u_{T-1}$, we have
$$
\int_{\mathbb{R}^m} \pi_{T-1}(u) \mathrm{d}u=1, \quad \pi_{T-1}(u)\geq 0 \quad a.e. \mbox{ on $\mathbb{R}^m$}. 
$$
Hence, 
\begin{equation*}\label{eq:pi_{T-1}}
\begin{aligned}
\pi_{T-1}^*(u)
& = \frac{\exp\Bigg\{-\dsp\frac{1}{\lambda}\left(2 \begin{pmatrix}
x_{T-1} \\ y_{T-1}\end{pmatrix}'H_{T-1} u + u' G_{T-1} u\right)\Bigg\}}{\dsp\int_{\mathbb{R}^m}\exp\Bigg\{-\dsp\frac{1}{\lambda}\left(2 \begin{pmatrix}
x_{T-1} \\ y_{T-1}\end{pmatrix}'H_{T-1} u + u' G_{T-1} u\right)\Bigg\}\mathrm{d}u} 
= \mathcal{N}\left(u\bigg\vert -G_{T-1}^{-1}H_{T-1}' \begin{pmatrix}x_{T-1} \\ y_{T-1}\end{pmatrix},~\frac{\lambda}{2}G_{T-1}^{-1}\right). 
\end{aligned}
\end{equation*}

Then a simple calculation for $\int_{\mathbb{R}^m} \pi_{T-1}(u) \ln \pi_{T-1}(u) \mathrm{d} u$ yields
$$
\begin{aligned}
& \int_{\mathbb{R}^m} \ln \pi_{T-1}(u) \pi_{T-1}(u) \mathrm{d}u \\
& = \int_{\mathbb{R}^m}\ln\left[{\bigg(\frac{1}{2\pi}\bigg)}^{\frac{m}{2}}{\left|\frac{\lambda}{2}G_{T-1}^{-1}\right|}^{-\frac{1}{2}}\exp \Bigg\{-\frac{1}{2}\left(u+G_{T-1}^{-1}H_{T-1}' \begin{pmatrix}x_{T-1}\\y_{T-1}\end{pmatrix}\right)'\frac{2}{\lambda}G_{T-1}\left(u+G_{T-1}^{-1}H_{T-1}' \begin{pmatrix}x_{T-1} \\ y_{T-1}\end{pmatrix}\right)\Bigg\}\right] \\
& \qquad \pi_{T-1}(u) \mathrm{d}u \\
& = \frac{1}{2}\ln \left[{\bigg(\frac{1}{\pi\lambda}\bigg)}^{m}|G_{T-1}|\right]-\frac{1}{\lambda}\mathbb{E}\left[u' G_{T-1} u\right]-\frac{2}{\lambda}\mathbb{E}\left[\begin{pmatrix}x_{T-1} \\ y_{T-1}\end{pmatrix}' H_{T-1} u\right] \\
& \quad -\frac{1}{\lambda}\begin{pmatrix}x_{T-1}\\y_{T-1}\end{pmatrix}' H_{T-1} G_{T-1}^{-1}H_{T-1}' \begin{pmatrix}x_{T-1} \\ y_{T-1}\end{pmatrix}.
\end{aligned}
$$
Substituting the above expression into  $J^\pi(T-1,x_{T-1},y_{T-1})$ leads to 
$$
\begin{aligned}
J^*\left(T-1,x_{T-1},y_{T-1}\right) 
& = \begin{pmatrix}x_{T-1}\\y_{T-1}\end{pmatrix}' F_{T-1} \begin{pmatrix}x_{T-1}\\y_{T-1}
\end{pmatrix}+\mathbb{E}\left[u_{T-1}' G_{T-1} u_{T-1}\right]+2\mathbb{E}\left[\begin{pmatrix}x_{T-1}\\y_{T-1}
\end{pmatrix}' H_{T-1} u_{T-1}\right] \\
& \quad + \frac{\lambda}{2}\ln \left[{\bigg(\frac{1}{\pi\lambda}\bigg)}^{m}|G_{T-1}|\right]-\mathbb{E}\left[u_{T-1}' G_{T-1} u_{T-1}\right]
   -2\mathbb{E}\left[\begin{pmatrix} x_{T-1}\\y_{T-1}\end{pmatrix}' H_{T-1} u_{T-1}\right] \\
& \quad -\begin{pmatrix}x_{T-1} \\ y_{T-1}\end{pmatrix}' H_{T-1} G_{T-1}^{-1}H_{T-1}' \begin{pmatrix}x_{T-1} \\ y_{T-1}\end{pmatrix} \\
\end{aligned}
$$
$$
\begin{aligned}
& = \begin{pmatrix}x_{T-1} \\ y_{T-1}\end{pmatrix}'\left(F_{T-1}-H_{T-1} G_{T-1}^{-1}H_{T-1}'\right)\begin{pmatrix}x_{T-1} \\ y_{T-1}\end{pmatrix}
      +\frac{\lambda}{2}\ln \left[{\bigg(\frac{1}{\pi\lambda}\bigg)}^m|G_{T-1}|\right] \\
& = \begin{pmatrix}x_{T-1} \\ y_{T-1}\end{pmatrix}' P_{T-1}\begin{pmatrix}x_{T-1} \\ y_{T-1}\end{pmatrix}
      +\frac{\lambda}{2}\ln \left[{\bigg(\frac{1}{\pi\lambda}\bigg)}^m|G_{T-1}|\right], 
\end{aligned}
$$
where $P_{T-1} = F_{T-1}-H_{T-1} G_{T-1}^{-1}H_{T-1}'$.

\vspace{0.5cm} 

Next, we derive the general form of $J^*(t,x_t,y_t)$ and $\pi^*_t(u)$, for $0 \leq t \leq T-1$. 

Assume that 
\begin{equation}
\begin{aligned}
J^*(t+1,x_{t+1},y_{t+1})
& = \begin{pmatrix}x_{t+1} \\ y_{t+1}\end{pmatrix}' P_{t+1}\begin{pmatrix}x_{t+1} \\ y_{t+1}\end{pmatrix} 
      + \frac{\lambda}{2}\sum_{k=t+1}^{T-1}\ln \left[{\bigg(\frac{1}{\pi\lambda}\bigg)}^m|G_{k}|\right] 
\end{aligned}	
\end{equation}
and 
\begin{equation}
\begin{aligned}
\pi_{t+1}^*(u)
& = \frac{\exp\Bigg\{-\dsp\frac{1}{\lambda}\left(2 \begin{pmatrix}x_{t+1}\\y_{t+1}\end{pmatrix}' H_{t+1} u+u' G_{t+1} u\right)\Bigg\}}{\dsp\int_{\mathbb{R}^m}\exp\Bigg\{-\frac{1}{\lambda}\left(2 \begin{pmatrix}x_{t+1} \\ y_{t+1}\end{pmatrix}' H_t u+u' G_{t+1} u\right)\Bigg\}\mathrm{d} u} 
= \mathcal{N}\left(u \bigg\vert -G_{t+1}^{-1}H_{t+1}'\begin{pmatrix}x_{t+1} \\ y_{t+1} \end{pmatrix},\frac{\lambda}{2}G_{t+1}^{-1}\right)
\end{aligned}
\end{equation}
hold at period $t+1$, where 
$$
\begin{aligned} 
P_{t+1} & = F_{t+1} - H_{t+1} G_{t+1}^{-1}H_{t+1}', \\ 
F_{t+1} & = \begin{pmatrix}A & 0 \\ 0 & \bar A\end{pmatrix}'P_{t+2}\begin{pmatrix}A & 0 \\ 0 & \bar A \end{pmatrix}
                   + \begin{pmatrix}C & 0 \\ 0 & \bar C \end{pmatrix}'P_{t+2}\begin{pmatrix}C & 0 \\ 0 & \bar C\end{pmatrix}, \\
H_{t+1} & = \begin{pmatrix}A & 0 \\ 0 & \bar A\end{pmatrix}'P_{t+2}\begin{pmatrix}B \\ 0\end{pmatrix}
                    + \begin{pmatrix}C & 0 \\0 & \rho\bar C\end{pmatrix}' P_{t+2}\begin{pmatrix} D \\ 0 \end{pmatrix}, \\
G_{t+1} & = \begin{pmatrix}B \\ 0\end{pmatrix}'P_{t+2}\begin{pmatrix}B \\ 0\end{pmatrix}
                    + \begin{pmatrix} D\\ 0 \end{pmatrix}'P_{t+2}\begin{pmatrix} D\\ 0 \end{pmatrix}. 
  \end{aligned}
$$
Then we consider 
$$
\begin{aligned}
J^\pi(t,x_t,y_t) 
& = \mathbb{E}\left[J^*(t+1,x_{t+1},y_{t+1}) + \lambda  \int_{\mathbb{R}^m} \pi_t(u) \ln \pi_t(u) \mathrm{d}u\Bigg| \mathcal{F}_{t} \right] \\
& = \mathbb{E}\left[\begin{pmatrix}x_{t+1} \\ y_{t+1}\end{pmatrix}' P_{t+1}\begin{pmatrix}x_{t+1} \\ y_{t+1}\end{pmatrix} 
      + \frac{\lambda}{2}\sum_{k=t+1}^{T-1}\ln \left[{\bigg(\frac{1}{\pi\lambda}\bigg)}^m|G_{k}|\right] 
      + \lambda  \int_{\mathbb{R}^m} \pi_t(u) \ln \pi_t(u) \mathrm{d}u \Bigg| \mathcal{F}_{t} \right] \\
& = \begin{pmatrix}x_{t} \\ y_{t} \end{pmatrix}'\left(\begin{pmatrix}A & 0 \\ 0 & \bar A\end{pmatrix}'P_{t+1}\begin{pmatrix}A & 0 \\ 0 & \bar A \end{pmatrix}
                   + \begin{pmatrix}C & 0 \\ 0 & \bar C \end{pmatrix}'P_{t+1}\begin{pmatrix}C & 0 \\ 0 & \bar C\end{pmatrix}\right)\begin{pmatrix}x_{t} \\ y_{t} \end{pmatrix} \\
& \quad + 2\int_{\mathbb{R}^m} \begin{pmatrix}x_{t} \\ y_{t} \end{pmatrix}'\left(\begin{pmatrix}A &0\\0&\bar A  \end{pmatrix}' P_{t+1} \begin{pmatrix}B \\0 \end{pmatrix}
      + \begin{pmatrix}C &0\\0&\rho\bar C\end{pmatrix}' P_{t+1} \begin{pmatrix}D \\0\end{pmatrix}\right)u_t \pi_t(u) \mathrm{d} u \\
& \quad + \int_{\mathbb{R}^m}\left[u'\left(\begin{pmatrix}B \\0 \end{pmatrix}' P_{t+1}\begin{pmatrix}B \\0 \end{pmatrix} 
   + \begin{pmatrix}D \\ 0\end{pmatrix}' P_{t+1} \begin{pmatrix}D \\0\end{pmatrix}\right)u\right]\pi_t(u) \mathrm{d}u \\
& \quad + \frac{\lambda}{2}\sum_{k=t+1}^{T-1}\ln \left[{\bigg(\frac{1}{\pi\lambda}\bigg)}^m|G_{k}|\right] + \lambda\int_{\mathbb{R}^m} \pi_t(u) \ln \pi_t(u) \mathrm{d}u \\
& = \begin{pmatrix}x_{t} \\ y_{t} \end{pmatrix}' F_t\begin{pmatrix}x_{t} \\ y_{t} \end{pmatrix} 
      + \frac{\lambda}{2}\sum_{k=t+1}^{T-1}\ln \left[{\bigg(\frac{1}{\pi\lambda}\bigg)}^m|G_{k}|\right] \\
& \quad + \int_{\mathbb{R}^m} \left[2 \begin{pmatrix}x_t \\ y_t\end{pmatrix}' H_t u +u' G_t u+\lambda \ln \pi_t(u)\right]\pi_t(u) \mathrm{d}u, 
\end{aligned}
$$
where
$$
\begin{aligned}
F_{t} & = \begin{pmatrix}A & 0 \\ 0 & \bar A\end{pmatrix}'P_{t+1}\begin{pmatrix}A & 0 \\ 0 & \bar A \end{pmatrix}
                   + \begin{pmatrix}C & 0 \\ 0 & \bar C \end{pmatrix}'P_{t+1}\begin{pmatrix}C & 0 \\ 0 & \bar C\end{pmatrix}, \\
H_{t} & = \begin{pmatrix}A & 0 \\ 0 & \bar A\end{pmatrix}'P_{t+1}\begin{pmatrix}B \\ 0\end{pmatrix}
                    + \begin{pmatrix}C & 0 \\0 & \rho\bar C\end{pmatrix}' P_{t+1}\begin{pmatrix} D \\ 0 \end{pmatrix}, \\
G_{t} & = \begin{pmatrix}B \\ 0\end{pmatrix}'P_{t+1}\begin{pmatrix}B \\ 0\end{pmatrix}
                    + \begin{pmatrix} D\\ 0 \end{pmatrix}'P_{t+1}\begin{pmatrix} D\\ 0 \end{pmatrix}. 
  \end{aligned}
$$

It follows from $\frac{\partial J^{\pi}(t,x_{t},y_{t})}{\partial \pi_{t}(u)} = 0$ that we have 
$$
2 \begin{pmatrix} x_{t} \\ y_{t} \end{pmatrix}' H_{t} u + u' G_{t} u + \lambda\ln \pi_{t}(u) + \lambda = 0.
$$
Since $\pi_{t}(u)$ is a probability density function of $u$ at period $t$, we have
$$
\int_{\mathbb{R}^m} \pi_{t}(u) \mathrm{d}u=1, \quad \pi_{t}(u)\geq 0 \quad a.e. \mbox{ on $\mathbb{R}^m$}. 
$$
Hence, 
\begin{equation*}\label{eq:pi_t}
\begin{aligned}
\pi_t^*(u)
& = \frac{\exp\Bigg\{-\dsp\frac{1}{\lambda}\left(2 \begin{pmatrix}
x_{t} \\ y_{t}\end{pmatrix} H_{t} u + u' G_{t}u\right)\Bigg\}}{\dsp\int_{\mathbb{R}^m}\exp\Bigg\{-\dsp\frac{1}{\lambda}\left(2 \begin{pmatrix}
x_{t} \\ y_{t}\end{pmatrix} H_{t} u + u' G_{t}u\right)\Bigg\}\mathrm{d}u} 
= \mathcal{N}\left(u \bigg\vert -G_{t}^{-1}H_{t}' \begin{pmatrix}x_{t} \\ y_{t}
\end{pmatrix},~\frac{\lambda}{2}G_{t}^{-1}\right). 
\end{aligned}
\end{equation*}

Then a simple calculation for $\int_{\mathbb{R}^m} \pi_{t}(u) \ln \pi_{t}(u) \mathrm{d} u$ yields
$$
\begin{aligned}
& \int_{\mathbb{R}^m} \ln \pi_{t}(u) \pi_{t}(u) \mathrm{d} u\\
& = \int_{\mathbb{R}^m}\ln\left[{\bigg(\frac{1}{2\pi}\bigg)}^{\frac{m}{2}}\left|\frac{\lambda}{2}G_{t}^{-1}\right|^{-\frac{1}{2}}\exp \Bigg\{-\frac{1}{2}\left(u+G_t^{-1}H_{t}' \begin{pmatrix}x_{t}\\y_{t}\end{pmatrix}\right)'\frac{2}{\lambda}G_{t}\left(u+G_t^{-1}H_{t}' \begin{pmatrix}x_{t}\\y_{t}\end{pmatrix}\right)\Bigg\}\right] \pi_{t}(u) \mathrm{d}u \\
& = \frac{\lambda}{2}\ln \left[{\bigg(\frac{1}{\pi\lambda}\bigg)}^m|G_{t}|\right]-\frac{1}{\lambda}\mathbb{E}\left[u' G_{t} u\right]-\frac{2}{\lambda}\mathbb{E}\left[\begin{pmatrix}x_{t} \\ y_{t}\end{pmatrix}' H_{t} u\right] - \frac{1}{\lambda}\begin{pmatrix}x_{t}\\y_{t}\end{pmatrix}' H_{t} G_{t}^{-1}H_{t}' \begin{pmatrix}x_{t} \\ y_{t}\end{pmatrix}.
\end{aligned}
$$
Substituting the above expression into  $J^\pi(t,x_{t},y_{t})$ leads to 
\begin{equation*}
\begin{aligned}
J^*(t,x_{t},y_{t}) 
& = \begin{pmatrix}x_{t}\\y_{t}\end{pmatrix}' F_{t} \begin{pmatrix}x_{t} \\ y_{t}\end{pmatrix} 
      + \frac{\lambda}{2}\sum_{k=t+1}^{T-1}\ln \left[{\bigg(\frac{1}{\pi\lambda}\bigg)}^m|G_{k}|\right] 
      + \mathbb{E}\left[u_{t}' G_{t} u_{t}\right]+2\mathbb{E}\left[\begin{pmatrix}x_{t} \\ y_{t}\end{pmatrix}' H_{t} u_{t}\right] \\
& \quad + \frac{\lambda}{2}\ln \left[{\bigg(\frac{1}{\pi\lambda}\bigg)}^m|G_{t}|\right]-\mathbb{E}\left[u_{t}' G_{t} u_{t}\right]
   - 2\mathbb{E}\left[\begin{pmatrix} x_{t}\\y_{t}\end{pmatrix}' H_{t} u_{t}\right] 
   - \begin{pmatrix}x_{t} \\ y_{t}\end{pmatrix}' H_{t} G_{t}^{-1}H_{t}' \begin{pmatrix}x_{t} \\ y_{t}\end{pmatrix} \\
& = \begin{pmatrix}x_{t} \\ y_{t}\end{pmatrix}'\left(F_{t}-H_{t} G_{t}^{-1}H_{t}'\right)\begin{pmatrix}x_{t} \\ y_{t}\end{pmatrix}
      + \frac{\lambda}{2}\sum_{k=t}^{T-1}\ln \left[{\bigg(\frac{1}{\pi\lambda}\bigg)}^m|G_{k}|\right] \\
& = \begin{pmatrix}x_{t} \\ y_{t}\end{pmatrix}' P_{t}\begin{pmatrix}x_{t} \\ y_{t}\end{pmatrix}
      + \frac{\lambda}{2}\sum_{k=t}^{T-1}\ln \left[{\bigg(\frac{1}{\pi\lambda}\bigg)}^m|G_{k}|\right], 
\end{aligned}
\end{equation*}
where $P_{t} = F_{t}-H_{t} G_{t}^{-1}H_{t}'$. 

This completes the proof. 
\hfill $\Box$

\vspace{0.5cm} 

\section{Proof of Theorem 2}
If $Q_T = \begin{pmatrix} 1 & -1 \\ -1 & 1\end{pmatrix}$, then $P_T = \begin{pmatrix} 1 & -1 \\ -1 & 1\end{pmatrix}$. 

At $t = T-1$, we have 
$$
\begin{aligned}
F_{T-1} & = \begin{pmatrix}A & 0 \\ 0 & \bar A\end{pmatrix}'P_{T}\begin{pmatrix}A & 0 \\ 0 & \bar A \end{pmatrix}
                   + \begin{pmatrix}C & 0 \\ 0 & \bar C \end{pmatrix}'P_{T}\begin{pmatrix}C & 0 \\ 0 & \bar C\end{pmatrix} 
                  = \begin{pmatrix}A^2 + C^2 & -A\bar A -C\bar C \\ -A\bar A -C\bar C & \bar A^2 + \bar C^2\end{pmatrix}, \\ 
H_{T-1} & = \begin{pmatrix}A & 0 \\ 0 & \bar A\end{pmatrix}'P_{T}\begin{pmatrix}B \\ 0\end{pmatrix}
                    + \begin{pmatrix}C & 0 \\0 & \rho\bar C\end{pmatrix}' P_{T}\begin{pmatrix} D \\ 0 \end{pmatrix} 
                    = \begin{pmatrix}AB + CD \\ -\bar AB -\rho\bar CD\end{pmatrix}, \\ 
G_{T-1} & = \begin{pmatrix}B \\ 0\end{pmatrix}'P_{T}\begin{pmatrix}B \\ 0\end{pmatrix}
                    + \begin{pmatrix} D\\ 0 \end{pmatrix}'P_{T}\begin{pmatrix} D\\ 0 \end{pmatrix} 
                 = B'B + D'D. 
\end{aligned}
$$ 
Hence, 
$$
\begin{aligned}
& P_{T-1} 
= F_{T-1}-H_{T-1} G_{T-1}^{-1}H_{T-1}' \\
& = \begin{pmatrix}A^2 + C^2 & -A\bar A - C\bar C\\ -A\bar A - C\bar C & \bar A^2 + \bar C^2\end{pmatrix}  
      - \begin{pmatrix}AB + CD \\ -\bar AB - \rho\bar CD\end{pmatrix}(B'B + D'D)^{-1}
      \begin{pmatrix}AB + CD \\ -\bar AB - \rho\bar CD\end{pmatrix}' \\ 
& {\small = \begin{pmatrix}A^2 + C^2 - (AB + CD)(B'B + D'D)^{-1}(AB + CD)' & -A\bar A - C\bar C + (AB + CD)(B'B + D'D)^{-1}(\bar AB + \rho\bar CD)'  \\ 
      -A\bar A - C\bar C + (AB + CD)(B'B + D'D)^{-1}(\bar AB + \rho\bar CD)' & \bar A^2 + \bar C^2 - (\bar AB + \rho\bar CD)(B'B + D'D)^{-1}(\bar AB + \rho\bar CD)' \end{pmatrix}. }
\end{aligned}
$$ 


At $t = T - 2$, we have 
$$
\begin{aligned}
F_{T-2} & = \begin{pmatrix}A & 0 \\ 0 & \bar A\end{pmatrix}'P_{T-1}\begin{pmatrix}A & 0 \\ 0 & \bar A \end{pmatrix}
                   + \begin{pmatrix}C & 0 \\ 0 & \bar C \end{pmatrix}'P_{T-1}\begin{pmatrix}C & 0 \\ 0 & \bar C\end{pmatrix} 
              = \begin{pmatrix}(A^2+C^2)P_{T-1,11} & (A\bar A+C\bar C)P_{T-1,12} \\ (A\bar A+C\bar C)P_{T-1,21} & (\bar A^2+\bar C^2)P_{T-1,22}\end{pmatrix}, \\ 
H_{T-2} & = \begin{pmatrix}A & 0 \\ 0 & \bar A\end{pmatrix}'P_{T-1}\begin{pmatrix}B \\ 0\end{pmatrix}
                    + \begin{pmatrix}C & 0 \\0 & \rho\bar C\end{pmatrix}' P_{T-1}\begin{pmatrix} D \\ 0 \end{pmatrix} 
                 = \begin{pmatrix}(AB+CD)P_{T-1,11} \\ (\bar AB+ \rho\bar CD)P_{T-1,21}\end{pmatrix}, \\ 
G_{T-2} & = \begin{pmatrix}B \\ 0\end{pmatrix}'P_{T-1}\begin{pmatrix}B \\ 0\end{pmatrix}
                    + \begin{pmatrix} D\\ 0 \end{pmatrix}'P_{T-1}\begin{pmatrix} D\\ 0 \end{pmatrix}
                 = (B'B + D'D)P_{T-1,11}.  
\end{aligned}
$$ 
Hence, 
$$
\begin{aligned}
& P_{T-2} 
= F_{T-2}-H_{T-2} G_{T-2}^{-1}H_{T-2}' \\
& = \begin{pmatrix}(A^2+C^2)P_{T-1,11} & (A\bar A+C\bar C)P_{T-1,12} \\ (A\bar A+C\bar C)P_{T-1,21} & (\bar A^2+\bar C^2)P_{T-1,22}\end{pmatrix} \\ 
& \quad - \begin{pmatrix}(AB+CD)P_{T-1,11} \\ (\bar AB+ \rho\bar CD)P_{T-1,21}\end{pmatrix}\left((B'B + D'D)P_{T-1,11}\right)^{-1}
   \begin{pmatrix}(AB+CD)P_{T-1,11} \\ (\bar AB+ \rho\bar CD)P_{T-1,21}\end{pmatrix}' \\ 
& {\scriptsize = \begin{pmatrix}
    \dsp\prod_{k=T-2}^{T-1}\!\!\left(A^2 + C^2 - (AB + CD)(B'B + D'D)^{-1}(AB + CD)'\right) 
    & -\dsp\prod_{k=T-2}^{T-1}\!\!\left(A\bar A + C\bar C - (AB + CD)(B'B + D'D)^{-1}(\bar AB + \rho\bar CD)'\right) \\ 
       -\dsp\!\!\prod_{k=T-2}^{T-1}\!\!\left(A\bar A + C\bar C - (AB + CD)(B'B + D'D)^{-1}(\bar AB + \rho\bar CD)'\right) & P_{T-2,22}\end{pmatrix}, } 
\end{aligned}
$$ 
where 
$$
\begin{aligned}
P_{T-2,22}
& = -\left(\dsp\sum_{j=T-2}^{T-1}\left(\prod_{k=j+1}^{T-1}\frac{(A\bar A + C\bar C - (\bar AB + \rho\bar C D)(B'B + D'D)^{-1}(AB + CD)')^2}{A^2 + C^2 - (AB + CD)(B'B + D'D)^{-1}(AB + CD)'}\right)\right. \\ 
& \quad \left.\times[(\bar AB + \rho\bar C D)(B'B + D'D)^{-1}(\bar AB + \rho\bar C D)']\left(\prod_{i=T-2}^{j-1}(\bar A^2 + \bar C^2)\right)\right) + (\bar A^2 + \bar C^2)^2. 
\end{aligned}
$$

At $t = T - 3$, we have 
$$
\begin{aligned}
F_{T-3} & = \begin{pmatrix}A & 0 \\ 0 & \bar A\end{pmatrix}'P_{T-2}\begin{pmatrix}A & 0 \\ 0 & \bar A \end{pmatrix}
                   + \begin{pmatrix}C & 0 \\ 0 & \bar C \end{pmatrix}'P_{T-2}\begin{pmatrix}C & 0 \\ 0 & \bar C\end{pmatrix} 
                  = \begin{pmatrix}(A^2+C^2)P_{T-2,11} & (A\bar A+C\bar C)P_{T-2,12} \\ 
                     (A\bar A+C\bar C)P_{T-2,21} & (\bar A^2+\bar C^2)P_{T-2,22}\end{pmatrix}, \\ 
H_{T-3} & = \begin{pmatrix}A & 0 \\ 0 & \bar A\end{pmatrix}'P_{T-2}\begin{pmatrix}B \\ 0\end{pmatrix}
                    + \begin{pmatrix}C & 0 \\0 & \rho\bar C\end{pmatrix}' P_{T-2}\begin{pmatrix} D \\ 0 \end{pmatrix} 
                 = \begin{pmatrix}(AB+CD)P_{T-2,11} \\ (\bar AB+ \rho\bar CD)P_{T-2,21}\end{pmatrix}, \\ 
G_{T-3} & = \begin{pmatrix}B \\ 0\end{pmatrix}'P_{T-2}\begin{pmatrix}B \\ 0\end{pmatrix}
                    + \begin{pmatrix} D\\ 0 \end{pmatrix}'P_{T-2}\begin{pmatrix} D\\ 0 \end{pmatrix}               
                    = (B'B + D'D)P_{T-2,11}.  
\end{aligned}
$$ 
Hence, 
$$
\begin{aligned}
& P_{T-3} 
= F_{T-3}-H_{T-3} G_{T-3}^{-1}H_{T-3}' \\
& = \begin{pmatrix}(A^2+C^2)P_{T-2,11} & (A\bar A+C\bar C)P_{T-1,12} \\ (A\bar A+C\bar C)P_{T-2,21} & (\bar A^2+\bar C^2)P_{T-2,22}\end{pmatrix} \\ 
& \quad - \begin{pmatrix}(AB+CD)P_{T-2,11} \\ (\bar AB+ \rho\bar CD)P_{T-2,21}\end{pmatrix}\left((B'B + D'D)P_{T-2,11}\right)^{-1}
   \begin{pmatrix}(AB+CD)P_{T-2,11} \\ (\bar AB+ \rho\bar CD)P_{T-2,21}\end{pmatrix}' \\ 
& {\scriptsize = \begin{pmatrix}
    \dsp\prod_{k=T-3}^{T-1}\!\!\left(A^2 + C^2 - (AB + CD)(B'B + D'D)^{-1}(AB + CD)'\right) 
    & -\dsp\!\!\prod_{k=T-3}^{T-1}\!\!\left(A\bar A + C\bar C - (AB + CD)(B'B + D'D)^{-1}(\bar AB + \rho\bar CD)'\right) \\ 
       -\dsp\!\!\prod_{k=T-3}^{T-1}\!\!\left(A\bar A + C\bar C - (AB + CD)(B'B + D'D)^{-1}(\bar AB + \rho\bar CD)'\right) & P_{T-3,22}\end{pmatrix}, } 
\end{aligned}
$$ 
where 
$$
\begin{aligned}
P_{T-3,22} 
& = -\left(\dsp\sum_{j=T-3}^{T-1}\left(\prod_{k=j+1}^{T-1}\frac{(A\bar A + C\bar C - (\bar AB + \rho\bar C D)(B'B + D'D)^{-1}(AB + CD)')^2}{A^2 + C^2 - (AB + CD)(B'B + D'D)^{-1}(AB + CD)' }\right)\right. \\ 
& \quad \left.\times[(\bar AB + \rho\bar C D)(B'B + D'D)^{-1}(\bar AB + \rho\bar C D)']\left(\prod_{i=T-3}^{j-1}(\bar A^2 + \bar C^2)\right)\right) + (\bar A^2 + \bar C^2)^3.  
\end{aligned}
$$

Therefore, for $t \leq T-1$, we similarly have 
$$
\begin{aligned}
F_{t} & = \begin{pmatrix}A & 0 \\ 0 & \bar A\end{pmatrix}'P_{t+1}\begin{pmatrix}A & 0 \\ 0 & \bar A \end{pmatrix}
                   + \begin{pmatrix}C & 0 \\ 0 & \bar C \end{pmatrix}'P_{t+1}\begin{pmatrix}C & 0 \\ 0 & \bar C\end{pmatrix} 
= \begin{pmatrix}(A^2+C^2)P_{t+1,11} & (A\bar A+C\bar C)P_{t+1,12} \\ 
                     (A\bar A+C\bar C)P_{t+1,21} & (\bar A^2+\bar C^2)P_{t+1,22}\end{pmatrix}, \\ 
H_{t} & = \begin{pmatrix}A & 0 \\ 0 & \bar A\end{pmatrix}'P_{t+1}\begin{pmatrix}B \\ 0\end{pmatrix}
                    + \begin{pmatrix}C & 0 \\ 0 & \rho\bar C\end{pmatrix}' P_{t+1}\begin{pmatrix} D \\ 0 \end{pmatrix} 
= \begin{pmatrix}(AB+CD)P_{t+1,11} \\ (\bar AB+ \rho\bar CD)P_{t+1,21}\end{pmatrix}, \\ 
G_{t} & = \begin{pmatrix}B \\ 0\end{pmatrix}'P_{t+1}\begin{pmatrix}B \\ 0\end{pmatrix}
                    + \begin{pmatrix} D\\ 0 \end{pmatrix}'P_{t+1}\begin{pmatrix} D\\ 0 \end{pmatrix} 
= (B'B + D'D)P_{t+1,11} \\ 
& = (B'B + D'D)\dsp\prod_{k=t+1}^{T-1}\left((A^2+C^2) - (AB+CD)(B'B + D'D)^{-1}(AB+CD)'\right) \\ 
& = (B'B + D'D)\left((A^2+C^2) - (AB+CD)(B'B + D'D)^{-1}(AB+CD)'\right)^{T-t-1}. 
\end{aligned}
$$ 
Hence, 
$$
\begin{aligned}
G_{t}^{-1}H_{t}' 
& = \left((B'B + D'D)P_{t+1,11}\right)^{-1}
   \begin{pmatrix}(AB+CD)P_{t+1,11} \\ (\bar AB+ \rho\bar CD)P_{t+1,21}\end{pmatrix}' \\ 
& = \begin{pmatrix}(B'B + D'D)^{-1}(AB+CD)' & & \dsp\frac{P_{t+1,21}}{P_{t+1,11}}(B'B + D'D)^{-1}(\bar AB+ \rho\bar CD)'\end{pmatrix} \\ 
& = \begin{pmatrix}(B'B + D'D)^{-1}(AB+CD)' & & g_t(B'B + D'D)^{-1}(\bar AB+ \rho\bar CD)'\end{pmatrix}, 
\end{aligned}
$$ 
where 
$$
\begin{aligned}
P_{t+1,11}
& = \dsp\prod_{k=t+1}^{T-1}\left((A^2+C^2) - (AB+CD)(B'B + D'D)^{-1}(AB+CD)'\right) \\ 
& = \left((A^2+C^2) - (AB+CD)(B'B + D'D)^{-1}(AB+CD)'\right)^{T-t-1}, \\ 
P_{t+1,21} & = P_{t+1,12} \\ 
& = -\dsp\prod_{k=t+1}^{T-1}\left(A\bar A + C\bar C - (AB + CD)(B'B + D'D)^{-1}(\bar AB + \rho\bar CD)' \right) \\ 
& = -\left(A\bar A + C\bar C - (AB + CD)(B'B + D'D)^{-1}(\bar AB + \rho\bar CD)' \right)^{T-t-1}, \\ 
g_t 
& = \frac{P_{t+1,21}}{P_{t+1,11}} 
= -\left(\frac{A\bar A + C\bar C - (AB + CD)(B'B + D'D)^{-1}(\bar AB + \rho\bar CD)'}{(A^2+C^2) - (AB+CD)(B'B + D'D)^{-1}(AB+CD)'}\right)^{T-t-1}. 
\end{aligned}
$$

Also, we have 
$$
\begin{aligned}
P_{t} 
& = F_{t}-H_{t} G_{t}^{-1}H_{t}' \\
& = \begin{pmatrix}(A^2+C^2)P_{t+1,11} & (A\bar A+C\bar C)P_{t+1,12} \\ (A\bar A+C\bar C)P_{t+1,21} & (\bar A^2+\bar C^2)P_{t+1,22}\end{pmatrix} \\ 
& \quad - \begin{pmatrix}(AB+CD)P_{t+1,11} \\ (\bar AB+ \rho\bar CD)P_{t+1,21}\end{pmatrix}\left((B'B + D'D)P_{t+1,11}\right)^{-1}
   \begin{pmatrix}(AB+CD)P_{t+1,11} \\ (\bar AB+ \rho\bar CD)P_{t+1,21}\end{pmatrix}' \\ 
& {\scriptsize = \begin{pmatrix}\dsp\prod_{k=t}^{T-1}\left((A^2+C^2) - (AB+CD)(B'B + D'D)^{-1}(AB+CD)'\right) 
                 & -\dsp\prod_{k=t}^{T-1}\left(A\bar A+C\bar C - (\bar AB + \rho\bar C D)(B'B + D'D)^{-1}(AB+CD)'\right) \\ 
                    -\dsp\prod_{k=t}^{T-1}\left(A\bar A+C\bar C - (\bar AB + \rho\bar C D)(B'B + D'D)^{-1}(AB+CD)'\right) & P_{t,22}\end{pmatrix} } \\ 
& {\scriptsize = \begin{pmatrix}\left((A^2+C^2) - (AB+CD)(B'B + D'D)^{-1}(AB+CD)'\right)^{T-t} 
                 & -\left(A\bar A+C\bar C - (\bar AB + \rho\bar C D)(B'B + D'D)^{-1}(AB+CD)'\right)^{T-t} \\ 
                    -\left(A\bar A+C\bar C - (\bar AB + \rho\bar C D)(B'B + D'D)^{-1}(AB+CD)'\right)^{T-t} & P_{t,22}\end{pmatrix}, } \\ 
\end{aligned}
$$ 
where 
$$
\begin{aligned}
P_{t,22} 
& = -\Bigg(\dsp\sum_{j=t}^{T-1}\left(\frac{(A\bar A + C\bar C - (\bar AB + \rho\bar C D)(B'B + D'D)^{-1}(AB + CD)')^2}{A^2 + C^2 - (AB + CD)(B'B + D'D)^{-1}(AB + CD)' }\right)^{T-j-1} \\ 
& \quad \times[(\bar AB + \rho\bar C D)(B'B + D'D)^{-1}(\bar AB + \rho\bar C D)'](\bar A^2 + \bar C^2)^{j-t}\Bigg) + (\bar A^2 + \bar C^2)^{T-t}. 
\end{aligned}
$$

Next, we consider 
$$
\begin{aligned}
& \frac{\lambda}{2}\sum_{k=t}^{T-1}\ln \left[{\bigg(\frac{1}{\pi\lambda}\bigg)}^m|G_{k}|\right] \\ 
& = \frac{\lambda}{2}\sum_{k=t}^{T-1}\ln \left[{\bigg(\frac{1}{\pi\lambda}\bigg)}^m\left|(B'B + D'D)\left((A^2+C^2) - (AB+CD)(B'B + D'D)^{-1}(AB+CD)'\right)^{T-k-1}\right|\right] \\ 
& = \frac{\lambda}{2}\ln\bigg(\frac{1}{\pi\lambda}\bigg)^m(T-t) + \frac{\lambda}{2}\ln\left|B'B + D'D\right|(T-t) \\ 
& \quad + \frac{\lambda}{4}\ln\left((A^2+C^2) - (AB+CD)(B'B + D'D)^{-1}(AB+CD)'\right)(T-t-1)(T-t). 
\end{aligned} 
$$
This completes the proof. 
\hfill $\Box$

\vspace{0.5cm} 

\section{Proof of Lemma \ref{lem:KLN}} 

If 
\begin{align*}
\bar\pi_{T-1}(u) 
& = \mathcal{N} \left(u \Big| K\begin{pmatrix}x_{T-1} \\ l_{T-1} \end{pmatrix}, \lambda LN^{T-(T-1)-1}\right) 
= \mathcal{N} \left(u \Big| K\begin{pmatrix}x_{T-1} \\ l_{T-1} \end{pmatrix}, \lambda L\right), 
\end{align*}
then 
\begin{align*}
& \E[u|\bar\pi_{T-1}] = K\begin{pmatrix}x_{T-1} \\ l_{T-1} \end{pmatrix}, \\
& \E[u^2|\bar\pi_{T-1}] = \begin{pmatrix}x_{T-1} \\ l_{T-1} \end{pmatrix}'K'K\begin{pmatrix}x_{T-1} \\ l_{T-1} \end{pmatrix} + \lambda L. 
\end{align*} 
Now, we first consider MV problem from period $T$ to $T-1$. The value function at period $T-1$ can be calculated as 
$$
\begin{aligned}
J^{\bar\pi}(T-1,x_{T-1},y_{T-1}) 
& = \mathbb{E}\left[\begin{pmatrix} x_{T} \\ y_{T} \end{pmatrix}' Q_T \begin{pmatrix} x_{T} \\ y_{T} \end{pmatrix}  
      + \lambda \int_{\mathbb{R}}\ln(\bar\pi_{T-1}(u))\bar\pi_{T-1}(u) \mathrm{d} u \bigg| \mathcal{F}_{T-1}\right] \\
& = \begin{pmatrix}
x_{T-1} \\
y_{T-1}
\end{pmatrix}'
\left[\begin{pmatrix}A & 0 \\ 0 & \bar A\end{pmatrix}'
Q_T
\begin{pmatrix}A & 0 \\ 0 & \bar A \end{pmatrix}
+ \begin{pmatrix}0 & 0 \\ 0 & \bar C \end{pmatrix}'
Q_T
\begin{pmatrix}0 & 0 \\ 0 & \bar C\end{pmatrix}\right]\begin{pmatrix}
x_{T-1} \\
y_{T-1}
\end{pmatrix} \\
& \quad + 2\dsp\int_{\mathbb{R}^m}\begin{pmatrix}
x_{T-1} \\
y_{T-1}
\end{pmatrix}'\left[\begin{pmatrix}
A & 0 \\ 0 & \bar A\end{pmatrix}'
Q_T
\begin{pmatrix}B \\ 0\end{pmatrix}
+ \begin{pmatrix}0 & 0 \\0 & \rho\bar C\end{pmatrix}' 
Q_T \begin{pmatrix} D \\ 0 \end{pmatrix}\right] 
u \bar\pi_{T-1}(u) \mathrm{d}u \\
& \quad + \dsp\int_{\mathbb{R}} \left[\begin{pmatrix}B \\ 0\end{pmatrix}'
Q_T
\begin{pmatrix}B \\ 0\end{pmatrix}
+ \begin{pmatrix} D\\ 0 \end{pmatrix}'
Q_T
\begin{pmatrix} D\\ 0 \end{pmatrix}\right]
u^2\bar\pi_{T-1}(u) \mathrm{d}u 
+ \lambda \int_{\mathbb{R}} \ln(\bar\pi_{T-1}(u))\bar\pi_{T-1}(u) \mathrm{d}u \\
& = \begin{pmatrix} x_{T-1} \\ y_{T-1} \end{pmatrix} F_{T-1} \begin{pmatrix} x_{T-1} \\ y_{T-1} \end{pmatrix} 
+ \int_{\mathbb{R}} \left[2 \begin{pmatrix} x_{T-1} \\ y_{T-1} \end{pmatrix}' H_{T-1} u + G_{T-1}u^2 
+ \lambda\ln\bar\pi_{T-1}(u)\right]\bar\pi_{T-1}(u) \mathrm{d}u,
\end{aligned}
$$
where
$$
\begin{aligned}
F_{T-1} & = \begin{pmatrix}A & 0 \\ 0 & \bar A\end{pmatrix}'Q_{T}\begin{pmatrix}A & 0 \\ 0 & \bar A \end{pmatrix}
                   + \begin{pmatrix}0 & 0 \\ 0 & \bar C \end{pmatrix}'Q_{T}\begin{pmatrix}0 & 0 \\ 0 & \bar C\end{pmatrix} 
                  = \begin{pmatrix}A^2 & -A\bar A \\ -A\bar A & \bar A^2 + \bar C^2\end{pmatrix}, \\ 
H_{T-1} & = \begin{pmatrix}A & 0 \\ 0 & \bar A\end{pmatrix}'Q_{T}\begin{pmatrix}B \\ 0\end{pmatrix}
                    + \begin{pmatrix}0 & 0 \\0 & \rho\bar C\end{pmatrix}' Q_{T}\begin{pmatrix} D \\ 0 \end{pmatrix} 
                    = \begin{pmatrix}AB \\ -\bar AB -\rho\bar CD\end{pmatrix}, \\ 
G_{T-1} & = \begin{pmatrix}B \\ 0\end{pmatrix}'Q_T\begin{pmatrix}B \\ 0\end{pmatrix}
                    + \begin{pmatrix} D\\ 0 \end{pmatrix}'Q_T\begin{pmatrix} D\\ 0 \end{pmatrix} 
                 = B^2 + D^2.
\end{aligned}
$$

Then a simple calculation for $\int_{\mathbb{R}} \ln(\bar\pi_{T-1}(u))\bar\pi_{T-1}(u) \mathrm{d}u$ yields
$$
\begin{aligned}
\int_{\mathbb{R}} \ln(\bar\pi_{T-1}(u))\bar\pi_{T-1}(u) \mathrm{d}u 
& = \int_{\mathbb{R}}\ln\left[\bigg(\frac{1}{\sqrt{2\pi}\sqrt{\lambda L}}\bigg)\exp \Bigg\{-\frac{1}{2\lambda B}\left(u - K\begin{pmatrix}x_{T-1} \\ l_{T-1}\end{pmatrix}\right)^2\Bigg\}\right]\bar\pi_{T-1}(u) \mathrm{d}u \\
& = \frac{1}{2}\ln\bigg(\frac{1}{2\pi\lambda L}\bigg)  - \frac{1}{2}. 
\end{aligned}
$$
Thus, we can get 
\begin{equation*}
\begin{aligned}
& J^{\bar\pi}\left(T-1,x_{T-1},l_{T-1}\right) \\
& = \begin{pmatrix} x_{T-1} \\ l_{T-1} \end{pmatrix} F_{T-1} \begin{pmatrix} x_{T-1} \\ l_{T-1} \end{pmatrix} 
+ \int_{\mathbb{R}} \left[2 \begin{pmatrix} x_{T-1} \\ l_{T-1} \end{pmatrix}' H_{T-1} u + G_{T-1}u^2 
+ \lambda\ln \bar\pi_{T-1}(u)\right] \bar\pi_{T-1}(u) \mathrm{d}u \\ 
& = \begin{pmatrix}x_{T-1} \\ l_{T-1}\end{pmatrix}' F_{T-1} \begin{pmatrix}x_{T-1} \\ l_{T-1}\end{pmatrix} 
+ 2\begin{pmatrix}x_{T-1} \\ l_{T-1}\end{pmatrix}' H_{T-1} \mathbb{E}[u|\bar\pi_{T-1}] 
+ G_{T-1}\mathbb{E}[u^2|\bar\pi_{T-1}] 
+ \frac{\lambda}{2}\ln\bigg(\frac{1}{2\pi\lambda L}\bigg) - \frac{\lambda}{2} \\ 
& = \begin{pmatrix}x_{T-1} \\ l_{T-1}\end{pmatrix}' F_{T-1} \begin{pmatrix}x_{T-1} \\ l_{T-1}\end{pmatrix} 
+ 2\begin{pmatrix}x_{T-1} \\ l_{T-1}\end{pmatrix}' H_{T-1} K\begin{pmatrix}x_{T-1} \\ l_{T-1} \end{pmatrix}
+ G_{T-1}\left(\begin{pmatrix}x_{T-1} \\ l_{T-1} \end{pmatrix}'K'K\begin{pmatrix}x_{T-1} \\ l_{T-1} \end{pmatrix} + \lambda L\right) \\
& \quad + \frac{\lambda}{2}\ln\bigg(\frac{1}{2\pi\lambda L}\bigg)  - \frac{\lambda}{2} \\ 
& = \begin{pmatrix}x_{T-1} \\ l_{T-1}\end{pmatrix}' M_{T-1} \begin{pmatrix}x_{T-1} \\ l_{T-1}\end{pmatrix} 
+ G_{T-1}\lambda L - \frac{\lambda}{2}\ln(2\pi\lambda L)  - \frac{\lambda}{2}, 
\end{aligned}
\end{equation*}
where 
$$
\begin{aligned} 
M_{T-1} 
& = F_{T-1} + 2H_{T-1}K + G_{T-1}K'K \\ 
& = \begin{pmatrix} A^2 + 2ABK_1 + (B^2+D^2)K_1'K_1 & & -A\bar A + 2ABK_2 + (B^2+D^2)K_1'K_2 \\
-A\bar A - 2(\bar AB + \rho\bar CD)K_1 + (B^2+D^2)K_2'K_1 & & \bar A^2 + \bar C^2 - 2(\bar AB + \rho\bar CD)K_2 + (B^2+D^2)K_2'K_2 \\ 
\end{pmatrix} \\ 
& = \begin{pmatrix} M_{T-1,11}^{} & M_{T-1,12} \\
M_{T-1,21} & M_{T-1,22} \\ 
\end{pmatrix} 
 = \begin{pmatrix} \bar m & M_{T-1,12} \\
M_{T-1,21} & M_{T-1,22} \\ 
\end{pmatrix} 
\end{aligned} 
$$ 

If 
\begin{align*}
\bar\pi_{T-2}(u) 
& = \mathcal{N} \left(u \Big| K\begin{pmatrix}x_{T-2} \\ l_{T-2} \end{pmatrix}, \lambda LN^{T-(T-2)-1}\right) 
= \mathcal{N} \left(u \Big| K\begin{pmatrix}x_{T-1} \\ l_{T-2} \end{pmatrix}, \lambda LN\right), 
\end{align*}
then 
\begin{align*}
& \E[u|\bar\pi_{T-2}] = K\begin{pmatrix}x_{T-2} \\ l_{T-2} \end{pmatrix}, \\
& \E[u^2|\bar\pi_{T-2}] = \begin{pmatrix}x_{T-2} \\ l_{T-2} \end{pmatrix}'K'K\begin{pmatrix}x_{T-2} \\ l_{T-2} \end{pmatrix} + \lambda LN. 
\end{align*} 
Now, we first consider MV problem from period $T-1$ to $T-2$. The value function at period $T-2$ can be calculated from the following equation
$$
\begin{aligned}
& J^{\bar\pi}(T-2,x_{T-2},y_{T-2}) \\
& = \mathbb{E}\left[\begin{pmatrix} x_{T-1} \\ y_{T-1} \end{pmatrix}' M_{T-1} \begin{pmatrix} x_{T-1} \\ y_{T-1} \end{pmatrix}  
      + \lambda \int_{\mathbb{R}^m} \ln(\bar\pi_{T-2}(u))\bar\pi_{T-2}(u) \mathrm{d} u \bigg| \mathcal{F}_{T-2}\right] 
      + G_{T-1}\lambda L - \frac{\lambda}{2}\ln(2\pi\lambda L)  - \frac{\lambda}{2} \\
& = \begin{pmatrix}
x_{T-2} \\
y_{T-2}
\end{pmatrix}'
\left[\begin{pmatrix}A & 0 \\ 0 & \bar A\end{pmatrix}'
M_{T-1}
\begin{pmatrix}A & 0 \\ 0 & \bar A \end{pmatrix}
+ \begin{pmatrix}0 & 0 \\ 0 & \bar C \end{pmatrix}'
M_{T-1}
\begin{pmatrix}0 & 0 \\ 0 & \bar C\end{pmatrix}\right]\begin{pmatrix}
x_{T-2} \\
y_{T-2}
\end{pmatrix} \\
& \quad + 2\dsp\int_{\mathbb{R}^m}\begin{pmatrix}
x_{T-2} \\
y_{T-2}
\end{pmatrix}'\left[\begin{pmatrix}
A & 0 \\ 0 & \bar A\end{pmatrix}'
M_{T-1}
\begin{pmatrix}B \\ 0\end{pmatrix}
+ \begin{pmatrix}0 & 0 \\0 & \rho\bar C\end{pmatrix}' 
M_{T-1} \begin{pmatrix} D \\ 0 \end{pmatrix}\right] 
u \bar\pi_{T-2}(u) \mathrm{d}u \\
& \quad + \dsp\int_{\mathbb{R}} \left[\begin{pmatrix}B \\ 0\end{pmatrix}'
M_{T-1}
\begin{pmatrix}B \\ 0\end{pmatrix}
+ \begin{pmatrix} D\\ 0 \end{pmatrix}'
M_{T-1}
\begin{pmatrix} D\\ 0 \end{pmatrix}\right]
u^2 \bar\pi_{T-2}(u) \mathrm{d}u \\ 
& \quad + \lambda \int_{\mathbb{R}} \bar\pi_{T-2}(u) \ln \bar\pi_{T-2}(u) \mathrm{d}u 
+ G_{T-1}\lambda L - \frac{\lambda}{2}\ln(2\pi\lambda L)  - \frac{\lambda}{2} \\
& = \begin{pmatrix} x_{T-2} \\ y_{T-2} \end{pmatrix} F_{T-2} \begin{pmatrix} x_{T-2} \\ y_{T-2} \end{pmatrix} 
+ \int_{\mathbb{R}} \left[2 \begin{pmatrix} x_{T-2} \\ y_{T-2} \end{pmatrix}' H_{T-2} u + G_{T-2}u^2 
+ \lambda\ln \bar\pi_{T-2}(u)\right] \bar\pi_{T-2}(u) \mathrm{d}u \\ 
& \quad + G_{T-1}\lambda L - \frac{\lambda}{2}\ln(2\pi\lambda L)  - \frac{\lambda}{2},
\end{aligned}
$$
where
$$
\begin{aligned}
F_{T-2} & = \begin{pmatrix}A & 0 \\ 0 & \bar A\end{pmatrix}'M_{T-1}\begin{pmatrix}A & 0 \\ 0 & \bar A \end{pmatrix}
                   + \begin{pmatrix}0 & 0 \\ 0 & \bar C \end{pmatrix}'M_{T-1}\begin{pmatrix}0 & 0 \\ 0 & \bar C\end{pmatrix} 
                  = \begin{pmatrix}A^2M_{T-1,11}^{} & A\bar AM_{T-1,12}^{} \\ A\bar AM_{T-1,21}^{} & (\bar A^2 + \bar C^2)M_{T-1,22}^{}\end{pmatrix}, \\ 
H_{T-2} & = \begin{pmatrix}A & 0 \\ 0 & \bar A\end{pmatrix}'M_{T-1}\begin{pmatrix}B \\ 0\end{pmatrix}
                    + \begin{pmatrix}0 & 0 \\0 & \rho\bar C\end{pmatrix}'M_{T-1}\begin{pmatrix} D \\ 0 \end{pmatrix} 
                    = \begin{pmatrix}ABM_{T-1,11}^{} \\ (\bar AB + \rho\bar CD)M_{T-1,21}^{}\end{pmatrix}, \\ 
G_{T-2} & = \begin{pmatrix}B \\ 0\end{pmatrix}'M_{T-1}\begin{pmatrix}B \\ 0\end{pmatrix}
                    + \begin{pmatrix} D\\ 0 \end{pmatrix}'M_{T-1}\begin{pmatrix} D\\ 0 \end{pmatrix} 
                 = (B^2 + D^2)M_{T-1,11}^{}. 
\end{aligned}
$$

Then a simple calculation for $\int_{\mathbb{R}} \ln(\bar\pi_{T-2}(u))\bar\pi_{T-2}(u) \mathrm{d}u$ yields
$$
\begin{aligned}
\int_{\mathbb{R}} \ln(\bar\pi_{T-2}(u))\bar\pi_{T-2}(u) \mathrm{d}u 
& = \int_{\mathbb{R}}\ln\left[\bigg(\frac{1}{\sqrt{2\pi}\sqrt{\lambda LN}}\bigg)\exp \Bigg\{-\frac{1}{2\lambda LN}\left(u - K\begin{pmatrix}x_{T-2} \\ l_{T-2}\end{pmatrix}\right)^2\Bigg\}\right]\bar\pi_{T-2}(u) \mathrm{d}u \\
& = \frac{1}{2}\ln\bigg(\frac{1}{2\pi\lambda LN}\bigg)  - \frac{1}{2}. 
\end{aligned}
$$
Thus, we can get 
\begin{equation*}
\begin{aligned}
& J^{\bar\pi}\left(T-2,x_{T-2},l_{T-2}\right) \\
& = \begin{pmatrix} x_{T-2} \\ l_{T-2} \end{pmatrix} F_{T-2} \begin{pmatrix} x_{T-2} \\ l_{T-2} \end{pmatrix} 
+ \int_{\mathbb{R}} \left[2 \begin{pmatrix} x_{T-2} \\ l_{T-2} \end{pmatrix}' H_{T-2} u + G_{T-2}u^2 
+ \lambda\ln\bar\pi_{T-2}(u)\right] \bar\pi_{T-2}(u) \mathrm{d}u \\ 
& \quad + G_{T-1}\lambda L - \frac{\lambda}{2}\ln(2\pi\lambda L)  - \frac{\lambda}{2} \\ 
& = \begin{pmatrix}x_{T-2} \\ l_{T-2}\end{pmatrix}' F_{T-2} \begin{pmatrix}x_{T-2} \\ l_{T-2}\end{pmatrix} 
+ 2\begin{pmatrix}x_{T-2} \\ l_{T-2}\end{pmatrix}' H_{T-2} \mathbb{E}[u|\bar\pi_{T-2}] 
+ G_{T-2}\mathbb{E}[u^2|\bar\pi_{T-2}] \\
& \quad + \frac{\lambda}{2}\ln\bigg(\frac{1}{2\pi\lambda LN}\bigg)  - \frac{1}{2LN}\left(\mathbb{E}[u^2|\pi_{T-2}^0] 
- 2\begin{pmatrix}x_{T-2} \\ l_{T-2}\end{pmatrix}'K'\mathbb{E}[u|\bar\pi_{T-2}] 
+ \begin{pmatrix}x_{T-2} \\ l_{T-2}\end{pmatrix}' K'K\begin{pmatrix}x_{T-2} \\ l_{T-2}\end{pmatrix}\right) \\ 
& \quad + G_{T-1}\lambda L - \frac{\lambda}{2}\ln(2\pi\lambda L)  - \frac{\lambda}{2} \\ 
& = \begin{pmatrix}x_{T-2} \\ l_{T-2}\end{pmatrix}' F_{T-2} \begin{pmatrix}x_{T-2} \\ l_{T-2}\end{pmatrix} 
+ 2\begin{pmatrix}x_{T-2} \\ l_{T-2}\end{pmatrix}' H_{T-2} K\begin{pmatrix}x_{T-2} \\ l_{T-2} \end{pmatrix}
+ G_{T-2}\left(\begin{pmatrix}x_{T-2} \\ l_{T-2} \end{pmatrix}'K'K\begin{pmatrix}x_{T-2} \\ l_{T-2} \end{pmatrix} + \lambda LN\right) \\
& \quad + \frac{\lambda}{2}\ln\bigg(\frac{1}{2\pi\lambda LN}\bigg)  - \frac{\lambda}{2}
+ G_{T-1}\lambda L - \frac{\lambda}{2}\ln(2\pi\lambda L)  - \frac{\lambda}{2} \\ 
& = \begin{pmatrix}x_{T-2} \\ l_{T-2}\end{pmatrix}' M_{T-2} \begin{pmatrix}x_{T-2} \\ l_{T-2}\end{pmatrix} 
+ G_{T-2}\lambda LN - \frac{\lambda}{2}\ln(2\pi\lambda LN)  - \frac{\lambda}{2} 
+ G_{T-1}\lambda L - \frac{\lambda}{2}\ln(2\pi\lambda L)  - \frac{\lambda}{2} \\ 
& = \begin{pmatrix}x_{T-2} \\ l_{T-2}\end{pmatrix}' M_{T-2} \begin{pmatrix}x_{T-2} \\ l_{T-2}\end{pmatrix} 
+ (B^2+D^2)\lambda L + (B^2+D^2)\bar m\lambda LN - \frac{\lambda}{2}\ln(2\pi\lambda L)\times 2  - \frac{\lambda}{2}\times 2 - \frac{\lambda}{2}\ln(N) \\ 
& = \begin{pmatrix}x_{T-2} \\ l_{T-2}\end{pmatrix}' M_{T-2} \begin{pmatrix}x_{T-2} \\ l_{T-2}\end{pmatrix} 
+ \frac{\lambda L(B^2+D^2)\left(1-(N\bar m)^{T-(T-2)}\right)}{1-N\bar m} \\
& - \frac{\lambda}{2}\ln(2\pi\lambda L)\times (T-(T-2))  - \frac{\lambda}{2}\times (T-(T-2)) - \frac{\lambda}{2}\ln(N)\frac{(T-(T-2)-1)(T-(T-2))}{2},  
\end{aligned}
\end{equation*}
where 
$$
\begin{aligned} 
M_{T-2} 
& = F_{T-2} + 2H_{T-2}K + G_{T-2}K'K 
= \begin{pmatrix} M_{T-2,11} & M_{T-2,12} \\
M_{T-2,21} & M_{T-2,22}
\end{pmatrix} 
= \begin{pmatrix} \bar m^2 & M_{T-2,12} \\
M_{T-2,21} & M_{T-2,22}
\end{pmatrix}, \\
M_{T-2,11} & = (A^2 + 2ABK_1 + (B^2+D^2)K_1'K_1)M_{T-1,11} = (A^2 + 2ABK_1 + (B^2+D^2)K_1'K_1)^2 = \bar m^2, \\ 
M_{T-2,12} & = A\bar AM_{T-1,12} + (2ABK_2 + (B^2+D^2)K_1'K_2)M_{T-1,11}, \\ 
M_{T-2,21} & = (A\bar A + 2(\bar AB + \rho\bar CD)K_1)M_{T-1,21} + (B^2+D^2)K_2'K_1M_{T-1,11}, \\ 
M_{T-2,22} & = (\bar A^2 + \bar C^2)M_{T-1,22}  + 2(\bar AB + \rho\bar CD)K_2M_{T-1,21} + (B^2+D^2)K_2'K_2M_{T-1,11}. 
\end{aligned} 
$$ 

Repeating the above procedure, we can get the following result. 

If 
\begin{align*}
\bar\pi_{t}(u) 
& = \mathcal{N} \left(u \Big| K\begin{pmatrix}x_{t} \\ l_{t} \end{pmatrix}, \lambda LN^{T-t-1}\right), 
\end{align*}
then 
\begin{align*}
& \E[u|\bar\pi_{t}] = K\begin{pmatrix}x_{t} \\ l_{t} \end{pmatrix}, \\
& \E[u^2|\bar\pi_{t}] = \begin{pmatrix}x_{t} \\ l_{t} \end{pmatrix}'K'K\begin{pmatrix}x_{t} \\ l_{t} \end{pmatrix} + \lambda LN^{T-t-1},  
\end{align*} 
and 
\begin{equation*}
\begin{aligned}
J^{\bar\pi}\left(t,x_{t},l_{t}\right) 
& = \begin{pmatrix}x_{t} \\ l_{t}\end{pmatrix}' M_{t} \begin{pmatrix}x_{t} \\ l_{t}\end{pmatrix} 
+ \frac{\lambda L(B^2+D^2)\left(1-(N\bar m)^{T-t}\right)}{1-N\bar m} \\
& \quad - \frac{\lambda}{2}\ln(2\pi\lambda L)(T-t)  - \frac{\lambda}{2}(T-t) - \frac{\lambda}{2}\ln(N)\frac{(T-t-1)(T-t)}{2} \\ 
& = \begin{pmatrix}x_{t} \\ l_{t}\end{pmatrix}' M_{t} \begin{pmatrix}x_{t} \\ l_{t}\end{pmatrix} 
+ f(t), 
\end{aligned}
\end{equation*}
where 
{\small 
$$
\begin{aligned} 
M_{t} 
& = \begin{pmatrix} (A^2 + 2ABK_1 + (B^2+D^2)K_1'K_1)M_{t+1,11} & A\bar AM_{t+1,12} + (2ABK_2 + (B^2+D^2)K_1'K_2)M_{t+1,11}^{} \\
(A\bar A + 2(\bar AB + \rho\bar CD)K_1)M_{t+1,21} + (B^2+D^2)K_2'K_1M_{t+1,11} & M_{t,22} \end{pmatrix} \\ 
& = \begin{pmatrix} M_{t,11} & M_{t,12} \\
M_{t,21} & M_{t,22}
\end{pmatrix} 
= \begin{pmatrix} \bar m^{T-t} & M_{t,12} \\
M_{t,21} & M_{t,22}
\end{pmatrix}, \\ \\ 
M_{t,22} & = (\bar A^2 + \bar C^2)M_{t+1,22} + 2(\bar AB + \rho\bar CD)K_2M_{t+1,21}^{} + (B^2+D^2)K_2'K_2M_{t+1,11}
\end{aligned} 
$$ 
} 
and 
$$
\begin{aligned}
f(t) 
& = \frac{\lambda L(B^2+D^2)\left(1-(N\bar m)^{T-t}\right)}{1-N\bar m} 
- \frac{\lambda}{2}\ln(2\pi\lambda L)(T-t)  - \frac{\lambda}{2}(T-t) - \frac{\lambda}{2}\ln(N)\frac{(T-t-1)(T-t)}{2}. 
\end{aligned}
$$
This completes the proof. 
\hfill $\Box$

\vspace{0.5cm} 

\section{Proof of Theorem \ref{thm:update}}

First of all, we let $M_t^t = M_t$, $\bar M_t^t = M_t^t$ and $J^{\pi^0}(t+1,x_{t+1},l_{t+1}) = J^{\bar\pi}(t+1,x_{t+1},l_{t+1})
= \begin{pmatrix}x_{t+1} \\ l_{t+1}\end{pmatrix}' \bar M_t^{t+1} \begin{pmatrix}x_{t+1} \\ l_{t+1}\end{pmatrix} + f(t+1)$. 
It follows from $J^{\pi^0}(t+1,x_{t+1},l_{t+1}) = \begin{pmatrix}x_{t+1} \\ l_{t+1}\end{pmatrix}' \bar M_t^{t+1} \begin{pmatrix}x_{t+1} \\ l_{t+1}\end{pmatrix} + f(t+1)$ that we get 
\begin{equation*}
\begin{aligned}
J^{\pi^0}(t,x_{t},l_{t}) 
& = \mathbb{E}\left[J^{\pi^0}(t+1,x_{t+1},l_{t+1}) + \lambda\int_{\mathbb{R}}\pi_t^0(u)\ln\pi_t^0(u) \bigg| \mathcal{F}_{t}\right] \\ 
& = \mathbb{E}\left[\begin{pmatrix}x_{t+1} \\ l_{t+1}\end{pmatrix}' \bar M_t^{t+1} \begin{pmatrix}x_{t+1} \\ l_{t+1}\end{pmatrix} 
+ \lambda\int_{\mathbb{R}}\pi_t^0(u)\ln\pi_t^0(u) \bigg| \mathcal{F}_{t}\right] + f(t+1) \\ 
& = \begin{pmatrix}
x_{t} \\
y_{t}
\end{pmatrix}'
\left[\begin{pmatrix}A & 0 \\ 0 & \bar A\end{pmatrix}'
\bar M_t^{t+1}
\begin{pmatrix}A & 0 \\ 0 & \bar A \end{pmatrix}
+ \begin{pmatrix}0 & 0 \\ 0 & \bar C \end{pmatrix}'
\bar M_t^{t+1}
\begin{pmatrix}0 & 0 \\ 0 & \bar C\end{pmatrix}\right]\begin{pmatrix}
x_{t} \\
y_{t}
\end{pmatrix} \\
& \quad + 2\dsp\int_{\mathbb{R}}\begin{pmatrix}
x_{t} \\
y_{t}
\end{pmatrix}'\left[\begin{pmatrix}
A & 0 \\ 0 & \bar A\end{pmatrix}'
\bar M_t^{t+1}
\begin{pmatrix}B \\ 0\end{pmatrix}
+ \begin{pmatrix}0 & 0 \\0 & \rho\bar C\end{pmatrix}' 
\bar M_t^{t+1} \begin{pmatrix} D \\ 0 \end{pmatrix}\right] 
u \pi_{t}^0(u) \mathrm{d}u \\
& \quad + \dsp\int_{\mathbb{R}} \left[\begin{pmatrix}B \\ 0\end{pmatrix}'
\bar M_t^{t+1}
\begin{pmatrix}B \\ 0\end{pmatrix}
+ \begin{pmatrix} D\\ 0 \end{pmatrix}'
\bar{M}^{t+1}_t
\begin{pmatrix} D\\ 0 \end{pmatrix}\right]
u^2 \pi_{t}^0(u) \mathrm{d}u 
+ \lambda \int_{\mathbb{R}} \pi_{t}^0(u) \ln \pi_{t}^0(u) \mathrm{d}u + f(t+1) \\
& = \begin{pmatrix} x_{t} \\ l_{t} \end{pmatrix} F_{t} \begin{pmatrix} x_{t} \\ l_{t} \end{pmatrix} 
+ \int_{\mathbb{R}} \left[2\begin{pmatrix} x_{t} \\ l_{t} \end{pmatrix}' H_{t} u + G_{t}u^2
+ \lambda\ln \pi_{t}^0(u)\right] \pi_{t}^0(u) \mathrm{d}u + f(t+1), 
\end{aligned}
\end{equation*}
where
$$
\begin{aligned}
F_{t} & = \begin{pmatrix}A & 0 \\ 0 & \bar A\end{pmatrix}'\bar M_t^{t+1}\begin{pmatrix}A & 0 \\ 0 & \bar A \end{pmatrix}
                   + \begin{pmatrix}0 & 0 \\ 0 & \bar C \end{pmatrix}'\bar M_t^{t+1}\begin{pmatrix}0 & 0 \\ 0 & \bar C\end{pmatrix}
            = \begin{pmatrix}A^2\bar M_{t,11}^{t+1} & A\bar A\bar M_{t,12}^{t+1} \\ 
               A\bar A\bar M_{t,21}^{t+1}& (\bar A^2 + \bar C^2)\bar M_{t,22}^{t+1}\end{pmatrix}, \\ 
H_{t} & = \begin{pmatrix}A & 0 \\ 0 & \bar A\end{pmatrix}'\bar M_t^{t+1} \begin{pmatrix}B \\ 0\end{pmatrix}
                    + \begin{pmatrix}0 & 0 \\0 & \rho\bar C\end{pmatrix}'\bar M_t^{t+1}\begin{pmatrix} D \\ 0 \end{pmatrix}
            = \begin{pmatrix}AB\bar M_{t,11}^{t+1} \\ (\bar AB + \rho\bar CD)\bar M_{t,21}^{t+1} \end{pmatrix}, \\
G_{t} & = \begin{pmatrix}B \\ 0\end{pmatrix}'\bar M_t^{t+1}\begin{pmatrix}B \\ 0\end{pmatrix}
                    + \begin{pmatrix} D\\ 0 \end{pmatrix}'\bar M_t^{t+1}\begin{pmatrix} D\\ 0 \end{pmatrix} 
             = (B^2 + D^2)\bar M_{t,11}^{t+1}. 
\end{aligned}
$$
We can take $\pi_t^1(u ; x_t, l_t)=\dsp\arg \min _{\pi_t^0(u)} J^{\pi^0}(t, x ; w)$ to make one iteration. Let $\frac{\partial J^{\pi^0}(t, x_t, l_t)}{\partial \pi_t^0(u)}=0$. Then we can get
$$
2\begin{pmatrix} x_{t} \\ y_{t} \end{pmatrix}' H_{t} u + G_{t}u^2+\lambda \ln \pi_t^0(u)+\lambda = 0.
$$
Applying the usual verification technique and using the fact that $\pi \in \mathcal{P}(\mathbb{R})$ if and only if
$$
\int_{\mathbb{R}} \pi_t^0(u) d u=1 \quad \mbox{ and } \quad \pi_t^0(u) \geq 0 \mbox{ a.e on } \mathbb{R}, 
$$
we obtain the feedback control $\pi_t^1(u ; x, w)$ whose density function is given by
\begin{equation*}\label{eq:pi_{T-1}}
\begin{aligned}
\pi_t^1(u ; x_{t}, l_{t})
& = \frac{\exp\Bigg\{-\dsp\frac{1}{\lambda}\left(2 \begin{pmatrix}
x_{t} \\ l_{t}\end{pmatrix}'H_{t} u + G_{t}u^2\right)\Bigg\}}{\dsp\int_{\mathbb{R}}\exp\Bigg\{-\dsp\frac{1}{\lambda}\left(2 \begin{pmatrix}
x_{t} \\ l_{t}\end{pmatrix}'H_{t} u + G_{t}u^2\right)\Bigg\}\mathrm{d} u} 
= \mathcal{N}\left(u \bigg\vert -G_{t}^{-1}H_{t}' \begin{pmatrix}x_{t} \\ l_{t}\end{pmatrix}, ~ \frac{\lambda}{2}G_{t}^{-1}\right). 
\end{aligned}
\end{equation*}
Then 
$$
\begin{aligned}
\int_{\mathbb{R}} \ln(\pi_{t}^1(u))\pi_{t}^1(u) \mathrm{d}u 
& = \int_{\mathbb{R}}\ln\left[\bigg(\frac{\sqrt{2G_t}}{\sqrt{2\pi}\sqrt{\lambda}}\bigg)\exp \Bigg\{-\frac{G_t}{\lambda}\left(u + G_{t}^{-1}H_{t}' \begin{pmatrix}x_{t} \\ l_{t}\end{pmatrix}\right)^2\Bigg\}\right]\pi_{t}^1(u) \mathrm{d}u \\
& = \int_{\mathbb{R}}\left(\frac{1}{2}\ln\bigg(\frac{G_t}{\pi\lambda}\bigg) - \frac{G_t}{\lambda}\left(u + G_{t}^{-1}H_{t}' \begin{pmatrix}x_{t} \\ l_{t}\end{pmatrix}\right)^2\right)\pi_{t}^1(u) \mathrm{d}u \\
& = \frac{1}{2}\ln\bigg(\frac{G_t}{\pi\lambda}\bigg) - \frac{1}{\lambda}\int_{\mathbb{R}} \left[G_{t}u^2 + 2\begin{pmatrix} x_{t} \\ l_{t} \end{pmatrix}' H_{t} u 
+ \begin{pmatrix} x_{t} \\ l_{t} \end{pmatrix}' H_{t}G_{t}^{-1}H_{t}'\begin{pmatrix} x_{t} \\ l_{t} \end{pmatrix}\right] \mathrm{d}u. 
\end{aligned}
$$
Thus, we can get 
$$
\begin{aligned}
J^{\pi^1}(t,x_{t},l_{t}) 
& = \begin{pmatrix} x_{t} \\ l_{t} \end{pmatrix} F_{t} \begin{pmatrix} x_{t} \\ l_{t} \end{pmatrix} 
- \begin{pmatrix} x_{t} \\ l_{t} \end{pmatrix}' H_{t}G_{t}^{-1}H_{t}'\begin{pmatrix} x_{t} \\ l_{t} \end{pmatrix}  
+ \frac{\lambda}{2}\ln\bigg(\frac{G_t}{\pi\lambda}\bigg) + f(t+1) \\ 
& = \begin{pmatrix} x_{t} \\ l_{t} \end{pmatrix} F_{t} \begin{pmatrix} x_{t} \\ l_{t} \end{pmatrix} 
- \begin{pmatrix} x_{t} \\ l_{t} \end{pmatrix}' H_{t}G_{t}^{-1}H_{t}'\begin{pmatrix} x_{t} \\ l_{t} \end{pmatrix}  
+ \frac{\lambda}{2}\ln\bigg(\frac{B^2+D^2}{\pi\lambda}\bigg) + \frac{\lambda}{2}\ln(\bar m)(T-t-1) + f(t+1)\\
& = \begin{pmatrix} x_{t} \\ l_{t} \end{pmatrix} M_t^{t+1} \begin{pmatrix} x_{t} \\ l_{t} \end{pmatrix} 
+ \frac{\lambda}{2}\ln\bigg(\frac{B^2+D^2}{\pi\lambda}\bigg) + \frac{\lambda}{2}\ln(\bar m)(T-t-1) + f(t+1), 
\end{aligned}
$$ 
where $M_t^{t+1} = F_t - H_tG_t^{-1}H_t'$. 

Also, we can get the following result
$$
J^{\pi^1}(t,x_{t},l_{t})  \leq J^{\pi^0}(t,x_{t},l_{t}). 
$$
After updating our policy for $j$ times, we can get
$$
\begin{aligned}
J^{\pi^j}(t,x_{t},l_{t})
& = \begin{pmatrix} x_{t} \\ l_{t} \end{pmatrix} M_t^{t+j} \begin{pmatrix} x_{t} \\ l_{t} \end{pmatrix} 
+ \frac{\lambda}{2}j\ln\bigg(\frac{B^2+D^2}{\pi\lambda}\bigg) + \frac{\lambda}{4}(j-1)j\ln\left(\frac{A^2D^2}{B^2+D^2}\right) \\ 
& \quad + \frac{\lambda}{2}j\ln(\bar m)(T-t-j) + f(t+j). 
\end{aligned}
$$ 

Next, if follows from $J^{\pi^j}(t+1,x_{t+1},l_{t+1}) = \begin{pmatrix}x_{t+1} \\ l_{t+1}\end{pmatrix}' \bar M_{t+j+1} \begin{pmatrix}x_{t+1} \\ l_{t+1}\end{pmatrix} + \frac{\lambda}{2}(j+1)\ln\bigg(\frac{B^2+D^2}{\pi\lambda}\bigg) + \frac{\lambda}{4}j(j+1)\ln\left(\frac{A^2D^2}{B^2+D^2}\right) + \frac{\lambda}{2}(j+1)\ln(\bar m)(T-t-1-j) + f(t+j+1)
$ and $\bar M_t^{t+j} = M_t^{t+j}$ that we derive the $(j+1)$-th iteration 
\begin{equation*}
\begin{aligned}
J^{\pi^j}(t,x_{t},l_{t}) 
& = \mathbb{E}\left[J^{\pi^j}\left(t+1,x_{t+1},l_{t+1}\right) + \lambda\int_{\mathbb{R}}\pi_t^j(u)\ln\pi_t^j(u) \bigg| \mathcal{F}_{t}\right] + \frac{\lambda}{2}(j+1)\ln\bigg(\frac{B^2+D^2}{\pi\lambda}\bigg)  \\ 
& \quad + \frac{\lambda}{4}j(j+1)\ln\left(\frac{A^2D^2}{B^2+D^2}\right)+ \frac{\lambda}{2}(j+1)\ln(\bar m)(T-t-1-j) + f(t+j+1) \\ 
& = \mathbb{E}\left[\begin{pmatrix}x_{t+1} \\ l_{t+1}\end{pmatrix}' \bar M_{t+j+1} \begin{pmatrix}x_{t+1} \\ l_{t+1}\end{pmatrix} 
+ \lambda\int_{\mathbb{R}}\pi_t^j(u)\ln\pi_t^j(u) \bigg| \mathcal{F}_{t}\right] + \frac{\lambda}{2}(j+1)\ln\bigg(\frac{B^2+D^2}{\pi\lambda}\bigg)  \\ 
& \quad + \frac{\lambda}{4}j(j+1)\ln\left(\frac{A^2D^2}{B^2+D^2}\right)+ \frac{\lambda}{2}(j+1)\ln(\bar m)(T-t-1-j) + f(t+j+1) \\ 
& = \begin{pmatrix}
x_{t} \\
y_{t}
\end{pmatrix}'
\left[\begin{pmatrix}A & 0 \\ 0 & \bar A\end{pmatrix}'
\bar M_t^{t+j+1}
\begin{pmatrix}A & 0 \\ 0 & \bar A \end{pmatrix}
+ \begin{pmatrix}0 & 0 \\ 0 & \bar C \end{pmatrix}'
\bar M_t^{t+j+1}
\begin{pmatrix}0 & 0 \\ 0 & \bar C\end{pmatrix}\right]\begin{pmatrix}
x_{t} \\
y_{t}
\end{pmatrix} \\
& \quad + 2\dsp\int_{\mathbb{R}}\begin{pmatrix}
x_{t} \\
y_{t}
\end{pmatrix}'\left[\begin{pmatrix}
A & 0 \\ 0 & \bar A\end{pmatrix}'
\bar M_t^{t+j+1}
\begin{pmatrix}B \\ 0\end{pmatrix}
+ \begin{pmatrix}0 & 0 \\0 & \rho\bar C\end{pmatrix}' 
M_{t+j+1} \begin{pmatrix} D \\ 0 \end{pmatrix}\right] 
u \pi_{t}^j(u) \mathrm{d}u \\
& \quad + \dsp\int_{\mathbb{R}} \left[\begin{pmatrix}B \\ 0\end{pmatrix}'
\bar M_t^{t+j+1}
\begin{pmatrix}B \\ 0\end{pmatrix}
+ \begin{pmatrix} D\\ 0 \end{pmatrix}'
\bar M_t^{t+j+1}
\begin{pmatrix} D\\ 0 \end{pmatrix}\right]
u^2 \pi_{t}^j(u) \mathrm{d}u \\ 
& \quad + \lambda \int_{\mathbb{R}}\ln(\pi_{t}^j(u))\pi_{t}^j(u) \mathrm{d}u + \frac{\lambda}{2}(j+1)\ln\bigg(\frac{B^2+D^2}{\pi\lambda}\bigg)  \\ 
& \quad + \frac{\lambda}{4}j(j+1)\ln\left(\frac{A^2D^2}{B^2+D^2}\right)+ \frac{\lambda}{2}(j+1)\ln(\bar m)(T-t-1-j) + f(t+j+1) \\ 
& = \begin{pmatrix} x_{t} \\ l_{t} \end{pmatrix} F_{t+j} \begin{pmatrix} x_{t} \\ l_{t} \end{pmatrix} 
+ \int_{\mathbb{R}} \left[2\begin{pmatrix} x_{t} \\ l_{t} \end{pmatrix}' H_{t+j} u + G_{t+j}u^2
+ \lambda\ln \pi_{t}^j(u)\right] \pi_{t}^j(u) \mathrm{d}u + \frac{\lambda}{2}(j+1)\ln\bigg(\frac{B^2+D^2}{\pi\lambda}\bigg)  \\ 
& \quad + \frac{\lambda}{4}j(j+1)\ln\left(\frac{A^2D^2}{B^2+D^2}\right)+ \frac{\lambda}{2}(j+1)\ln(\bar m)(T-t-1-j) + f(t+j+1), 
\end{aligned}
\end{equation*}
where
$$
\begin{aligned}
F_{t+j} & = \begin{pmatrix}A & 0 \\ 0 & \bar A\end{pmatrix}'\bar M_t^{t+j+1}\begin{pmatrix}A & 0 \\ 0 & \bar A \end{pmatrix}
                   + \begin{pmatrix}0 & 0 \\ 0 & \bar C \end{pmatrix}'\bar M_t^{t+j+1}\begin{pmatrix}0 & 0 \\ 0 & \bar C\end{pmatrix}
            = \begin{pmatrix}A^2 \bar M_{t,11}^{t+j+1} & A\bar A\bar M_{t,12}^{t+j+1}  \\ 
               A\bar A\bar M_{t,21}^{t+j+1}  & (\bar A^2 + \bar C^2)\bar M_{t,22}^{t+j+1} \end{pmatrix}, \\ 
H_{t+j} & = \begin{pmatrix}A & 0 \\ 0 & \bar A\end{pmatrix}'\bar M_t^{t+j+1}\begin{pmatrix}B \\ 0\end{pmatrix}
                    + \begin{pmatrix}0 & 0 \\0 & \rho\bar C\end{pmatrix}'\bar M_t^{t+j+1}\begin{pmatrix} D \\ 0 \end{pmatrix}
            = \begin{pmatrix}AB\bar M_{t,11}^{t+j+1}  \\ (\bar AB + \rho\bar CD)\bar M_{t,21}^{t+j+1}  \end{pmatrix}, \\
G_{t+j} & = \begin{pmatrix}B \\ 0\end{pmatrix}'\bar M_t^{t+j+1}\begin{pmatrix}B \\ 0\end{pmatrix}
                    + \begin{pmatrix} D\\ 0 \end{pmatrix}'\bar M_t^{t+j+1}\begin{pmatrix} D\\ 0 \end{pmatrix} 
             = (B^2 + D^2) \bar M_{t,11}^{t+j+1}. 
\end{aligned}
$$
We can take $\pi_t^{j+1}(u ; x_t, l_t)=\dsp\arg \min _{\pi_t^j(u)} J^{\pi^j}(t, x ; w)$ to make one iteration. Let $\frac{\partial J^{\pi^j}(t, x_t, l_t)}{\partial \pi_t^j(u)}=0$. Then we can get
$$
2\begin{pmatrix} x_{t} \\ y_{t} \end{pmatrix}' H_{t+j} u + G_{t+j}u^2+\lambda \ln \pi_t^j(u)+\lambda=0. 
$$
Applying the usual verification technique and using the fact that $\pi \in \mathcal{P}(\mathbb{R})$ if and only if
$$
\int_{\mathbb{R}} \pi_t^j(u) d u=1 \quad \mbox{ and } \quad \pi_t^j(u) \geq 0 \mbox{ a.e on } \mathbb{R}, 
$$
we obtain the feedback control $\pi_t^{j+1}(u ; x, w)$ whose density function is given by
\begin{equation*}\label{eq:pi_{T-1}}
\begin{aligned}
\pi_t^{j+1}(u ; x_{t}, l_{t})
& = \frac{\exp\Bigg\{-\dsp\frac{1}{\lambda}\left(2 \begin{pmatrix}
x_{t} \\ l_{t}\end{pmatrix}'H_{t+j} u + G_{t+j}u^2\right)\Bigg\}}{\dsp\int_{\mathbb{R}}\exp\Bigg\{-\dsp\frac{1}{\lambda}\left(2 \begin{pmatrix}
x_{t} \\ l_{t}\end{pmatrix}'H_{t+j} u + G_{t+j}u^2\right)\Bigg\}\mathrm{d} u} 
= \mathcal{N}\left(u \bigg\vert -G_{t+j}^{-1}H_{t+j}' \begin{pmatrix}x_{t} \\ l_{t}\end{pmatrix}, ~ \frac{\lambda}{2}G_{t+j}^{-1}\right). 
\end{aligned}
\end{equation*}
Then 
$$
\begin{aligned}
\int_{\mathbb{R}} \ln \pi_{t}^{j+1}(u) \pi_{t}^{j+1}(u) \mathrm{d}u 
& = \int_{\mathbb{R}}\ln\left[\bigg(\frac{\sqrt{2G_{t+j}}}{\sqrt{2\pi}\sqrt{\lambda}}\bigg)\exp \Bigg\{-\frac{G_{t+j}}{\lambda}\left(u + G_{t+j}^{-1}H_{t+j}' \begin{pmatrix}x_{t} \\ l_{t}\end{pmatrix}\right)^2\Bigg\}\right]\pi_{t}^{j+1}(u) \mathrm{d}u \\
& = \int_{\mathbb{R}}\left(\frac{1}{2}\ln\bigg(\frac{G_{t+j}}{\pi\lambda}\bigg) - \frac{G_{t+j}}{\lambda}\left(u + G_{t+j}^{-1}H_{t+j}' \begin{pmatrix}x_{t} \\ l_{t}\end{pmatrix}\right)^2\right)\pi_{t}^{j+1}(u) \mathrm{d}u \\
& = \frac{1}{2}\ln\bigg(\frac{G_{t+j}}{\pi\lambda}\bigg) 
- \frac{1}{\lambda}\int_{\mathbb{R}} \left[G_{t+j}u^2 + 2\begin{pmatrix} x_{t} \\ l_{t} \end{pmatrix}' H_{t+j} u 
+ \begin{pmatrix} x_{t} \\ l_{t} \end{pmatrix}' H_{t+j}G_{t+j}^{-1}H_{t+j}'\begin{pmatrix} x_{t} \\ l_{t} \end{pmatrix}\right] \mathrm{d}u. 
\end{aligned}
$$						
Thus, we can get 
$$
\begin{aligned}
J^{\pi^{j+1}}(t,x_{t},l_{t}) 
& = \begin{pmatrix} x_{t} \\ l_{t} \end{pmatrix} F_{t+j} \begin{pmatrix} x_{t} \\ l_{t} \end{pmatrix} 
- \begin{pmatrix} x_{t} \\ l_{t} \end{pmatrix}' H_{t+j}G_{t+j}^{-1}H_{t+j}'\begin{pmatrix} x_{t} \\ l_{t} \end{pmatrix}  
+ \frac{\lambda}{2}j\ln\bigg(\frac{B^2+D^2}{\pi\lambda}\bigg) + \frac{\lambda}{4}(j-1)j\ln\left(\frac{A^2D^2}{B^2+D^2}\right) \\ 
& \quad + \frac{\lambda}{2}j\ln(\bar m)(T-t-1-j) + \frac{\lambda}{2}\ln\bigg(\frac{G_{t+j}}{\pi\lambda}\bigg) + f(t+j+1) \\ 
& = \begin{pmatrix} x_{t} \\ l_{t} \end{pmatrix} M_t^{t+j+1} \begin{pmatrix} x_{t} \\ l_{t} \end{pmatrix} 
+ \frac{\lambda}{2}(j+1)\ln\bigg(\frac{B^2+D^2}{\pi\lambda}\bigg) + \frac{\lambda}{4}j(j+1)\ln\left(\frac{A^2D^2}{B^2+D^2}\right) \\ 
& \quad + \frac{\lambda}{2}(j+1)\ln(\bar m)(T-t-1-j) + f(t+j+1), 
\end{aligned} 
$$ 
where $M_t^{t+j+1}  = F_{t+j} - H_{t+j}G_{t+j}^{-1}H_{t+j}'$. 

Also, we can get the following result
$$
J^{\pi^{j+1}}(t,x_{t},l_{t})  \leq J^{\pi^j}(t,x_{t},l_{t}). 
$$
In fact, after deriving $T-t$ iterations, $\pi_t^{T-t}(u ; x_{t}, l_{t})$ converges to $\pi_t^*(u ; x_{t}, l_{t})$ 
\begin{equation*}\label{eq:pi_{T-1}}
\begin{aligned}
\pi_t^{T-t}(u ; x_{t}, l_{t})
& = \mathcal{N}\left(u \bigg\vert -(G_t^*)^{-1}H_t^{*'} \begin{pmatrix} x_t\\y_t\end{pmatrix},~\frac{\lambda}{2}(G_t^*)^{-1}\right) 
= \pi_t^*(u), 
\end{aligned}
\end{equation*}
where
$$
\begin{aligned}
G_{t}^* 
& = (B^2 + D^2)\left(\frac{A^2D^2}{B^2 + D^2}\right)^{T-t-1}, \\ 
(G_{t}^*)^{-1}H_{t}^{*'} 
& = \begin{pmatrix}\dsp\frac{AB}{B^2 + D^2} & \quad & \dsp -\left(\frac{\bar AD - \rho B\bar C}{AD}\right)^{T-t-1}\frac{\bar AB+ \rho\bar CD}{B^2 + D^2}\end{pmatrix}, 
\end{aligned}
$$ 
for $t=0,1,\cdots T-1$. 

Using $f(t+T-t) = f(T) = 0$, we can prove that $J^{\pi^{T-t}}(t,x_t,l_t)$ converges to $J^*(t,x_t,l_t)$ 
$$
\begin{aligned}
J^{\pi^{T-t}}(t,x_t,l_t) 
& = \begin{pmatrix}x_t \\ l_t\end{pmatrix}' P_t \begin{pmatrix}x_t \\ l_t\end{pmatrix}
      + \frac{\lambda}{2}\ln\bigg(\frac{1}{\pi\lambda}\bigg)(T-t) + \frac{\lambda}{2}\ln\left(B^2 + D^2\right)(T-t) \\ 
& \quad + \frac{\lambda}{4}\ln\left(\frac{A^2D^2}{B^2+D^2}\right)(T-t-1)(T-t) + f(T) \\ 
& = \begin{pmatrix}x_t \\ l_t\end{pmatrix}' P_t \begin{pmatrix}x_t \\ l_t\end{pmatrix}
      + \frac{\lambda}{2}\ln\bigg(\frac{1}{\pi\lambda}\bigg)(T-t) + \frac{\lambda}{2}\ln\left(B^2 + D^2\right)(T-t) \\ 
& \quad + \frac{\lambda}{4}\ln\left(\frac{A^2D^2}{B^2+D^2}\right)(T-t-1)(T-t) \\ 
& = J^*(t,x_t,l_t). 
\end{aligned}
$$
This completes the proof. 
\hfill $\Box$

\end{document}